\documentclass[journal]{IEEEtran} 
\usepackage{url}
\usepackage{graphicx}
\usepackage{cite}

\usepackage{graphics} 
\usepackage{epsfig} 
\usepackage{mathptmx} 
\usepackage{amsmath} 
\usepackage{amssymb} 
\usepackage[ruled,vlined,linesnumbered]{algorithm2e}
\usepackage{hyperref}
\usepackage{subfigure}
\usepackage{newtxtext,newtxmath}
\usepackage{bm}
\usepackage{tabularx}
\usepackage{booktabs}
\usepackage{multirow}
\usepackage{float}
\usepackage{array}
\usepackage{mathtools}
\usepackage{tablefootnote}
\usepackage{xcolor}

\usepackage{fancyhdr} 
\usepackage{draftwatermark} 

\renewcommand{\headrulewidth}{0pt}
\makeatletter
\def\ps@IEEEtitlepagestyle{%
  \def\@oddfoot{\normalfont\hfil\thepage\hfil}%
  \def\@evenfoot{\normalfont\hfil\thepage\hfil}%
  \def\@oddhead{}%
  \def\@evenhead{}%
  \renewcommand{\headrulewidth}{0pt}%
  \renewcommand{\footrulewidth}{0pt}%
}

\def\maketitle{\par
  \begingroup
    \renewcommand\thefootnote{}
    \def\footnotemark{}
    \def\footnoterule{\relax}
    \if@twocolumn
      \twocolumn[\@maketitle]
      \else
      \@maketitle
    \fi
    \@thanks
  \endgroup
  \setcounter{footnote}{0}
  \let\maketitle\relax
  \let\@maketitle\relax
  \gdef\@thanks{}\gdef\@author{}\gdef\@title{}\gdef\@IEEEcompsoctitleabstractindextext{}%
  \def\@IEEEcompsoctitleabstractindextext{}%
  \def\@IEEEcompsoctitleabstractindextextfootnotetext{}%
  \def\@IEEEcompsoctitleabstractindextextfootnotetextfootnotetext{}%
  \thispagestyle{fancy}
}

\SetWatermarkText{Accepted Version}
\SetWatermarkScale{0.5}
\SetWatermarkColor[gray]{0.85}

\begin{document}

\title{Exploiting Kinematic Redundancy for\\Robotic Grasping of Multiple Objects}

\author{Kunpeng Yao,~\IEEEmembership{Member,~IEEE} and Aude Billard,~\IEEEmembership{Fellow,~IEEE}
\thanks{The authors are with the Learning Algorithms and Systems Laboratory (LASA), \'{E}cole Polytechnique F\'{e}d\'{e}rale de Lausanne (EPFL), CH-1015 Lausanne, Switzerland.
Corresponding author: Kunpeng Yao, \url{kunpeng.yao@epfl.ch}}
\thanks{This work was supported through an ERC Advanced Grant, project ID: 741945, funded by the European Commission.}}

\markboth{IEEE Transactions on Robotics}
{Shell \MakeLowercase{\textit{et al.}}: A Sample Article Using IEEEtran.cls for IEEE Journals}


\maketitle

\begin{abstract}
Humans coordinate the abundant degrees of freedom (DoFs) of hands to dexterously perform tasks in everyday life. We imitate human strategies to advance the dexterity of multi-DoF robotic hands. Specifically, we enable a robot hand to grasp multiple objects by exploiting its kinematic redundancy, referring to all its controllable DoFs.
We propose a human-like grasp synthesis algorithm to generate grasps using pairwise contacts on arbitrary opposing hand surface regions, no longer limited to fingertips or hand inner surface. To model the available space of the hand for grasp, we construct a reachability map, consisting of reachable spaces of all finger phalanges and the palm. It guides the formulation of a constrained optimization problem, solving for feasible and stable grasps. We formulate an iterative process to empower robotic hands to grasp multiple objects in sequence. Moreover, we propose a kinematic efficiency metric and an associated strategy to facilitate exploiting kinematic redundancy. We validated our approaches by generating grasps of single and multiple objects using various hand surface regions. Such grasps can be successfully replicated on a real robotic hand.
\end{abstract}


\section{Introduction}
\begin{figure}[htp!]
\centering
    \includegraphics[width=\columnwidth]{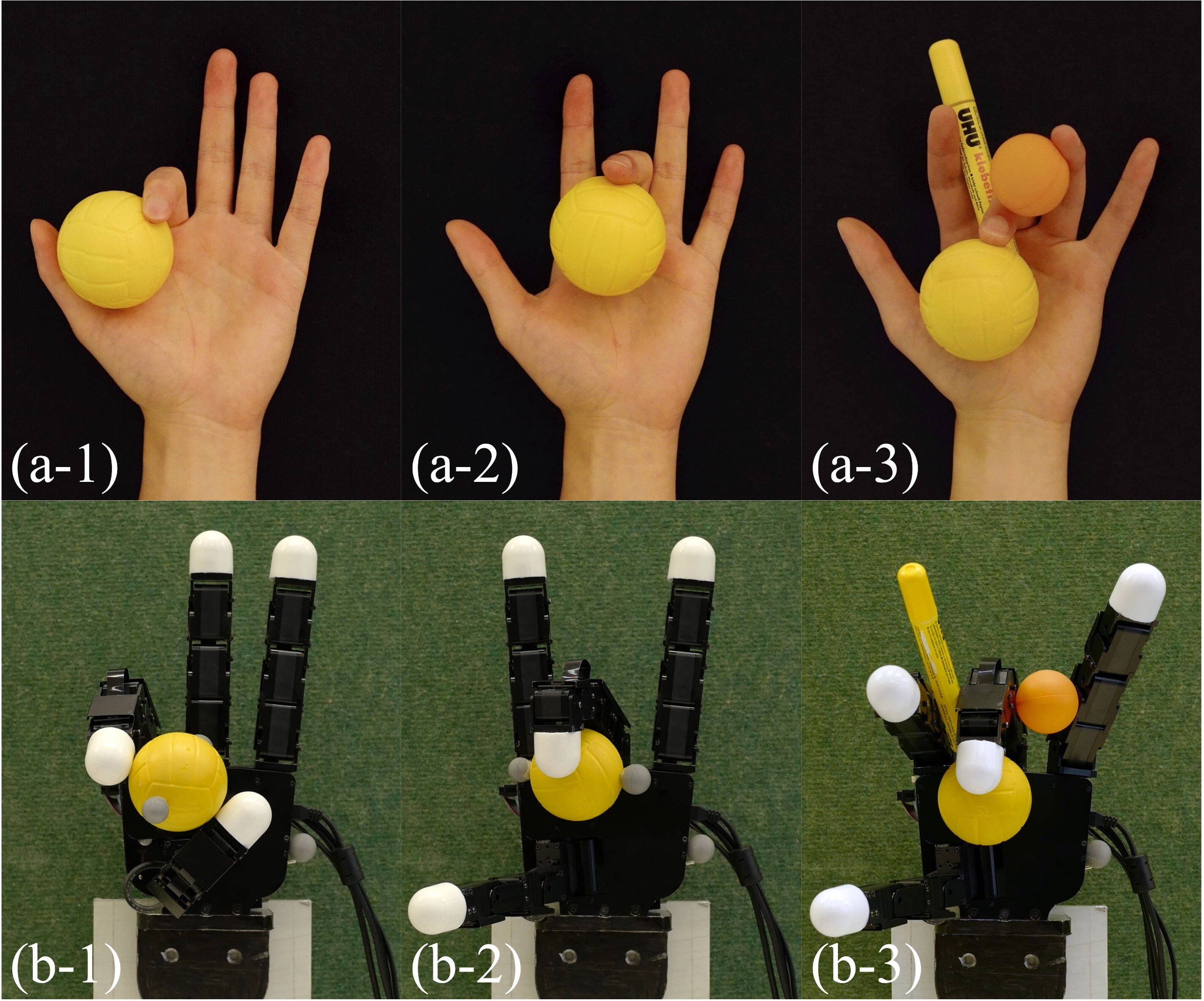}
    \caption{A multi-fingered robotic hand performs human-like dexterous grasps enabled by our proposed algorithms.
    (a-1) A human hand pinches a softball.
    (b-1) A robotic hand imitates the grasp in (a-1) using the same fingers.
    (a-2) A human hand wraps a softball.
    (b-2) A robotic hand imitates (a-2) to wrap the object.
    (a-3) A human hand grasps multiple objects.
    (b-3) A robotic hand imitates (a-3) to grasp multiple objects.}
    \label{fig:cover}
\end{figure}

\IEEEPARstart{G}{rasping} is a primitive but central skill for robotic hands and manipulators to interact with the environment.
It is also a prerequisite for any desired manipulation movement.
Grasping requires the hand to coordinate its controllable degrees of freedom to establish multiple contacts on the surface of the target object, to form a stable and collision-free configuration.
The problem of generating such a grasp configuration is known as \emph{grasp synthesis} \cite{shimoga1996robot}.
Despite recent advances in robotic grasping and in the design of complex robotic hands with sensing and actuation functionality, the community remains largely focused on grasping a \emph{single} object, albeit with either a \emph{power grasp}, i.e., using the inner surface of the hand to wrap the object \cite{zhuang2019shared}, or through a \emph{pinch grasp} using the thumb and the index finger in coordination \cite{deng2020grasping}.
Although such simple hand poses enable robotic hands to handle a majority of objects, it lacks the ability to dexterously employ the abundant DoFs of the hand in a grasp.
In comparison, humans can grasp objects with a variety of hand poses (see Fig.~\ref{fig:cover}), using not only the fingertips and the palm, but also almost any regions on the hand surface, as documented in multiple taxonomies of human hand poses \cite{gonzalez2014analysis,feix2015grasp,starke2020human}.
In this respect, the dexterity of robotic hands is still far from the dexterity of human hands \cite{billard2019trends}.

In addition, although many hands have abundant DoFs in their structure, most advanced robotic hands including industrial robotic manipulators are only capable of grasping one single object at a time; and even the local adjustment of fingers on the grasped object remains a challenging problem \cite{sundaralingam2018geometric}.
The robotic grasping of multiple objects has rarely been studied.
Some representative works have focused on the use of a simple robotic manipulator to envelope multiple objects at the same time, and the physical interaction among the objects enveloped as a whole \cite{harada1998kinematics,yamada2011grasp,yamada2015static}.
Human hands, in contrast, can grasp multiple objects simultaneously in everyday tasks.
For instance, holding different cutlery when clearing up dishes, or manipulating a pair of chopsticks to pick up food.
These types of grasps hold objects within the lengths of adjacent fingers \cite{feix2015grasp}.
While such grasps that use all the surfaces of fingers are common for humans, they are rarely seen in robotics that tend to favor fingertip grips.
Empowering robotic hands and manipulators with human-like dexterity will enable them to multitask, improving the efficiency of their task performance.

We address the above-mentioned limitations in current robotic grasping studies.
We propose a framework to enable a robotic hand to achieve human-like dexterity, such as grasping objects using arbitrary hand poses and grasping multiple objects at the same time.
We tackle this challenge in three steps.
First, we formulate the grasping problem in a generalized case.
Inspired by the concept of \emph{virtual finger}, we parameterize contacts, such that they can exist in arbitrary surface regions of the hand -- not only fingertips but also the lateral and back surfaces of any finger phalanges, also the palm surface. This allows us to use any part of the robotic hand to grasp the target object.
To validate our algorithm, we select a set of everyday objects, generate unconventional grasps in simulation (e.g., wrapping an object with one finger, or pinching an object with the lateral surfaces of fingers), and then reproduce them on a real robotic hand.
Secondly, based on our single-object grasping algorithm, we formulate an iterative process to enable the robotic hand to grasp multiple objects.
This is achieved by exploiting the robotic hand's \emph{kinematic redundancy}, defined as the set of remaining controllable degrees of freedom in a robot model that can be used for task planning.
Given multiple target objects, the robotic hand plans the grasp of each object in sequence.
Each time an object is grasped, the hand's DoFs used in grasping are removed; then, the hand grasps the next object using its kinematic redundancy.
In each experimental trial, we randomly select multiple objects, generate the grasp sequence in simulation, and then replicate the sequence step by step on a real robot.
Finally, to maximize the use of kinematic redundancy, we propose a \emph{kinematic efficiency metric}, which quantifies the occupied space and employed DoFs taken by each grasp.
By optimizing this metric at every step of a sequential grasp, the robotic hand prioritizes the grasp that absorbs the least possible kinematic redundancy, yielding a highly ``kinematic-efficient'' grasp.
Experimental results demonstrate that by applying this criterion, the robot hand can spare more kinematic redundancy at each step and manage to grasp more objects than without optimizing the kinematic efficiency metric.
This work will not only help to improve the dexterity of robotic hands and manipulators in grasping tasks, such as pick-and-place using multi-fingered grippers on industrial assembly lines, but will also inspire the development of algorithms for dexterous manipulation, and the design of dexterous robotic hands and manipulators.
This work's contributions are three folds:
\begin{itemize}
    \item We propose a human-like dexterous grasp synthesis algorithm that empowers a multi-DoF robotic hand to employ the opposition spaces spanned by hand links for grasping objects using arbitrary surface regions.
    \item We propose an iterative process that allows the robotic hand to grasp and hold multiple objects in sequence.
    \item We propose a kinematic efficiency metric and an associated strategy to facilitate the exploitation of kinematic redundancy in the hand model in sequential grasping.
\end{itemize}

\section{Related Work}
\subsection{Robot grasp synthesis}
The robot grasping problem has been extensively and intensively studied in the past decades \cite{bicchi2000robotic}.
Methods used include optimization-based approaches \cite{panagiotopoulos1994robot}, learning from human demonstration \cite{aleotti2010interactive}, statistical learning-based approaches \cite{murali2018learning}, and deep learning \cite{caldera2018review,andrychowicz2020learning}, which has become increasingly popular in recent years.
\cite{sahbani2012overview} summarizes recent advances in robot grasping algorithms, while \cite{ozawa2017grasp} provides a review of related studies from a control perspective.
Given the model of the robot hand and the model of the object to grasp, the grasp synthesis problem aims to find a feasible configuration of the hand-object system, by determining the contact points both on the hand surface and on the object surface.
The reach-and-grasp movements of humans typically consist of two steps \cite{rosenbaum2001planning}: the hand is pre-positioned near the object based on their relative spatial relationship (high-level planning), and then the hand moves its fingers to establish contact with the target object (low-level planning).
Such a two-stage grasping process is commonly used for planning robotic grasping and manipulation tasks, and we hence divide our review accordingly.

\subsubsection{High-level planning}
The high-level planning determines the relative spatial pose between the hand and the target object to grasp.
\cite{ciocarlie2009hand} introduced the concept of \emph{eigen grasp} to describe hand configurations in a low-dimensional subspace, based on the hand postural synergy \cite{santello1998postural}.
This is an object-centered approach, and the optimal grasping position of the hand with respect to the target object is determined by generating a large number of samples of the hand sub-spaces around the target object.
The kinematic properties of the robot hand model can be used to facilitate grasp positioning.
For example, \cite{zacharias2007capturing} sampled a robotic arm model in joint space and constructed a \emph{capability map} to describe the spatial reachability of the arm.
Spatial regions with high reachability index values are prioritized for task planning, such as generating manipulation sequences \cite{ruehl2011graspability}.
The capability map can be constructed offline and easily adapt to new tasks and objects.
Object models have also been exploited in grasp synthesis.
\cite{gienger2008task} introduced an object-specific \emph{task map} that represents a manifold of feasible power grasps on the target object and defined the goal as a hypervolume on the map.
This is useful to generate a feasible path for all fingers to close simultaneously to envelop the object.
It, however, does not consider other types of grasps than power grasp, nor the planning of individual finger positioning.
\cite{zacharias2009object} proposed an object-specific \emph{grasp map} to represent the characteristics of the target object, and showed that the combination of object grasp map and robot capability map in high-level planning greatly speeds up the low-level planning with guaranteed grasping quality.
\cite{roa2011graspability} constructed a \emph{graspability map} by combining a sampling-based approach with the hand's reachability map.
It employs the reachability map to restrict the spatial location of the hand relative to the object, and then samples the positions and orientations of the hand around the object in a discretized spatial mesh grid.
Only poses that result in collision-free and force-closure grasps are retained.
The optimal grasp is then selected from a set of feasible hand poses.
Afterward, the robot hand moves to this desired optimal position, adjusts its posture, and then closes its fingers to grasp.
The graspability map has demonstrated its efficiency in grasping tasks \cite{siciliano2012advanced} and has also guided the design of grippers \cite{eizicovits2016integration}.
However, the accuracy of the map depends largely on the voxelization of the spatial volume around the object. Resolution is often non-uniform and hence, at the same spatial location, the quality of the grasp may vary significantly for different orientations of the hand.
It is also time-consuming to evaluate the collision and stability conditions for all samples, especially since most of them end up being discarded.
Moreover, the graspability map must be recreated for any novel target objects.
Although it is possible to generate an approximation of the graspability map using shape primitives \cite{eizicovits2018automatic}, quality degradation is inevitable.
To handle these issues, we construct a hand-centered \emph{reachability map} by sampling in joint space, consisting of reachable spaces of each individual link of the hand (i.e., a finger phalanx or the palm), represented by three-dimensional alpha shapes \cite{edelsbrunner1994three}.
We use this reachability map to estimate a boundary for the spatial location of the target object relative to the hand.
The grasp configuration is then computed based on the geometric property of this reachability map.

\subsubsection{Low-level planning}
The low-level grasp planning problem generates a feasible configuration of the hand-object system.
This problem can be solved using either \emph{analytic approaches}, e.g., relying on solving a constrained optimization problem over grasp quality metrics; or \emph{empirical approaches} (i.e., data-driven approaches), by sampling numerous feasible configurations and selecting the optimal ones \cite{bohg2013data}.
Although empirical approaches do not require accurate model representations and afford direct sampling on real robotic hands, they demand a large number of samples in general.
Moreover, they are essentially \emph{object-centered} approaches \cite{goldfeder2011data} and can hardly be applied to generate grasps of multiple objects at once.
In this paper, we use an analytic approach by formulating the grasp synthesis as a constrained optimization problem \cite{el2013generation,el2015computing}, aiming at determining a configuration of the hand-object system, which is both collision-free and stable.
A number of metrics based on grasp configurations have been widely used in grasp synthesis \cite{mishra1995grasp,miller1999examples,borst2004grasp,roa2015grasp} and human grasp analysis \cite{leon2012evaluation}.
\cite{li1988task} tackled the optimal grasping problem by optimizing task-oriented quality metrics for multi-fingered robotic hands.
\cite{ponce1995computing} computed stable grasps of 2D polygonal objects using a three-fingered robotic hand.
Using geometric conditions for closure properties, \cite{zhu2004planning} proposed an algorithm for computing form/force closure grasps of 3D objects with curved surfaces and multiple contact points.
\cite{garcia2015dexterous} generated stable grasps of circular objects through two contacts.
More recently, \cite{el2015computing} solved the optimal grasping problem for a multi-fingered robotic hand in the form of joint space optimization, which is subject to collision-free and force-closure properties.
A comprehensive review of grasp synthesis algorithms can be found at \cite{shimoga1996robot,sahbani2012overview}.
Most works by default assume that contacts locate only at the fingertips and exclude other surface regions of the hand.
In power grasps, contacts are not the result of a selective choice of which surface to use.
\cite{sommer2016multi} offered an approach to selectively maximize contact regions along both the frontal and back surfaces of the fingers, to increase the number of contacts on the object.
However, the location of these contacts was pre-specified.
Although fingertips have higher motion freedom than other parts of the hand, using only fingertips cannot fulfill task-specified demands for all grasps \cite{el2013generation}.
In contrast, we provide a human-like solution that dexterously exploits full-hand kinematic structure to compensate for the insufficiency of fingertips.
We allow contacts to present over the entire finger surface and at any region on the palm without pre-assignment and further limitations.

\subsection{Grasping multiple objects}
While grasping a single object has been intensively studied, the problem of grasping multiple objects with a single hand has received much less attention.
In an early attempt, \cite{harada1998kinematics} tackled the problem of grasping two objects using a robotic manipulator.
Given the grasping model, this study analyzed the kinematics and internal forces required to stabilize the two target objects when grasping them simultaneously.
In a similar scenario, \cite{yamada2011grasp,yamada2013grasp} analyzed the stability condition for grasping multiple three-dimensional objects based on the grasp potential energy, using a grasp stiffness matrix \cite{yamada2009grasp}.
Other considerations related to multi-object grasping, such as the feasibility of placing fingertips \cite{yu2013analysis} and the analysis of contact surface geometry model \cite{yamada2015static} have also been discussed.
The above studies use only a specific grasp type, i.e., the \emph{envelope grasp}.
Multiple target objects are viewed as a whole, and are arranged either next to one another in a chain-like manner, or piled up into a pyramid shape. The fingers make contact with the two objects only at the outer of the chain or pile.
The objects lying in the middle are stabilized through contacts with their neighbor objects.
There again, points of force application are modeled on the fingertips that are in contact with objects on either side of the chain.
On the one hand, considering the interaction among multiple objects enveloped inside the hand largely increases the difficulty of the grasp synthesis.
On the other hand, when multiple objects are grasped, it is difficult to manipulate each object individually, because stabilizing all objects largely constrains the forces that the manipulator can apply to each object separately.
The problem of grasping and manipulating multiple objects has also been discussed in other scenarios.
For example, \cite{donald2000distributed} proposed an algorithm for moving multiple objects as a whole in a multi-robot system that is able to perform specific manipulation operations, but precisely grasping and manipulating every single object is rarely possible.
In a recent study, a novel robotic manipulator has been designed to grasp and handle multiple objects \cite{mucchiani2020novel}.
However, this highly relies on specific robotic devices and is thus difficult to generalize to other robotic hands or manipulators.

In this work, we consider a broader scenario of grasping multiple objects, whereby each object can be held by arbitrary hand surface regions and in various poses.
This potentially enables the manipulation of each one of the grasped objects through contacts and also avoids tackling the interaction among multiple objects.
We use the opposition spaces spanned by opposing hand surface regions to reformulate the problem of grasping multiple objects at a sequence of single-object grasping problems.
This sub-problem can be solved by our proposed dexterous grasp synthesis algorithm.

\subsection{Virtual finger and opposition space}
Studies on human hand poses in grasping have provided insights to help understand human dexterity, suggesting that the shaping of the hand plays a significant role in dexterous skills, such as playing piano \cite{furuya2011hand} or watchmaking craftsmanship \cite{yao2020inverse,yao2021hand}.
Human dexterity features the unique ability to coordinate efficiently the abundant DoFs in hand.
To interpret how humans organize and control the coordination of the fingers in a grasp, \cite{arbib1985coordinated} introduced the concept of \emph{virtual finger}.
A virtual finger represents a group of one or multiple real human fingers or even parts of the hand (e.g. the palm) that move in unison to achieve an independent function, such as applying force in a desired direction.
This concept provides a functional point of view of the hand that considers controlled DoFs per group. It associates this group to a function, distinguishing the primary functions that contribute to the task and secondary functions that do not directly contribute and hence can be ignored.
The concept of opposition space \cite{iberall1986opposition} expands this concept and explains how virtual fingers come to coordinate placement and forces to stabilize the grasp or generate desired motions on the object.
Opposition space can be used to determine the intended actions on the object when observing how humans place their fingers on objects \cite{biegstraaten2006relation,de2015recognizing,smeets2019review}.
An \emph{opposition space} (OS, plural OSes) is defined as the sub-spaces inside the hand model that are structured by opposing regions on virtual fingers, so opposing forces can be applied inside this space.
It has been proposed that three basic oppositions constitute human prehension \cite{iberall1997human}, i.e., (1) \emph{pad opposition}, formed by two finger pads opposing each other, (2) \emph{palm opposition}, structured by the palm of the hand and the finger patch opposing it, and (3) \emph{side opposition}, constructed by the lateral patches of fingers.
Most grasping types discussed in robotics are restricted to pad opposition (e.g., pinch grasp) and palm opposition (e.g., power grasp), while grasps using side opposition are rarely discussed.

We propose a dexterous grasp synthesis algorithm, inspired by the concepts of virtual finger and opposition space.
We enable contacts on arbitrary parts of the hand and map the kinematic chain from the hand base to the contact point onto a virtual finger.
Thus, only the DoFs affecting the contacts are considered in the current grasping.
Irrelevant DoFs are ignored and considered as \emph{kinematic redundancy} that can be exploited in future tasks.
In our modelling, any pair of opposing regions on the hand surface can generate an opposition space.
These spaces can be used to construct grasps of each object separately, in such a way as to ensure force-closure property.

\section{Notation and Models}\label{sec:modelling}
\subsection{Notation}
Table~\ref{tab:notation} lists important notations in this paper.
A spatial vector pointing from $\bm{a}\in\mathbf{R}^{3}$ to $\bm{b}\in\mathbf{R}^{3}$ is denoted as $\overrightarrow{\bm{a}\bm{b}}$;
$\bm{d}(\boldsymbol{\cdot},~\boldsymbol{\cdot})$ indicates the Euclidean distance; $\{\cdot\}$ denotes a set, $\vert\cdot\vert$ is the cardinality of the element, $\lVert\cdot\rVert$ is the Euclidean norm, and
$\langle \cdot,~\cdot \rangle$ is the inner product of quaternions.

\begin{table} [!t]
\centering\sf\small
\caption{List of notations.}\label{tab:notation}
\begin{tabularx}{0.45\textwidth}{l<:X}
\toprule
    \multicolumn{2}{l}{\textsc{Reference frames}} \\
    $\{\mathbb{H}\}$ & hand reference frame \\
    $\{\mathbb{O}\}$ & object reference frame \\
    $\{\mathbb{C}_{i}\}$ & local reference frame of the $i$th contact \\
    $\{\mathbb{L}_{i}\}$ & local reference frame of the $i$th link \\
    \midrule
    \multicolumn{2}{l}{\textsc{Models and sets}} \\
    $\mathcal{H}$ & hand model \\
    $\mathcal{O}$ & object model \\
    $\mathcal{K}$ & kinematic redundancy set \\
    $\mathcal{R}_{i}$ & the reachable space of the $i$th link \\
    $\{\mathcal{R}\}$ & the reachability map of the hand \\
    $\mathcal{M}_{\mathcal{C}}$ & the collision map of the hand \\
    $\mathcal{S}_{i,j}$ & the opposition space spanned by $\mathcal{R}_{i}$ and $\mathcal{R}_{j}$ \\
    $\mathcal{L}_{i}$ & the $i$th generalized link (a finger phalanx or a palm) \\
    \midrule
    \multicolumn{2}{l}{\textsc{Indices}} \\
    ${N}_{l}$ & total number of generalized links \\
    ${N}_{j}$ & total number of joints \\
    ${N}_{c}$ & total number of contacts \\
    ${N}_{o}$ & total number of objects \\
    \midrule
    \multicolumn{2}{l}{\textsc{Parameters and variables}} \\
    $\mathcal{Q}$ & grasp quality metric \\
    ${C}_{\mathcal{S}}$ & capacity of an opposition space $\mathcal{S}$ \\
    $\bm{o}$ & position of object center represented in $\{\mathbb{H}\}$ \\
    $\bm{q}^{\mathcal{O}}$ &
    orientation quaternion of object $\mathcal{O}$ in $\{\mathbb{H}\}$ \\
    $\bm{p}_{i}$ & the $i$th contact position on the hand in $\{\mathbb{H}\}$ \\
    $\bm{p}^{\ast}_{i}$ & projection of $\bm{p}_{i}$ on the object surface along $\overrightarrow{\bm{p}_{i}\bm{o}}$ \\
    $\rho_{i}$ & radical distance associated with $\bm{p}_{i}$\\
    $\phi_{i}$ & angular coordinate associated with $\bm{p}_{i}$ \\
    $\alpha_{i}$ & height ratio associated with $\bm{p}_{i}$ \\
    $\bm{n}_{i}$ & contact normal of the $i$th contact on the hand \\
    $\bm{f}^{i}$ & force applied on the $i$th contact \\
    $\mathcal{F}^{i}$ & the approximated friction cone on the $i$th contact \\
    $\mathbf{f}^{i}_{n}$ & the $n$th edge of $\mathcal{F}^{i}$ \\
    $\bm{d}^{i}$ & the vector from $\bm{o}$ to $\bm{p}_{i}$ \\
    $\mathbf{\tau}^{i}_{n}$ & torque generated on the object center by $\mathbf{f}^{i}_{n}$ \\
    $\bm{w}^{i}_{n}$ &
    primitive wrench associated with $\mathbf{f}^{i}_{n}$\\
    $\varphi^{i}_{n}$ & the positive coefficient associated with $\bm{w}^{i}_{n}$ \\
    $d$ & the grasp distance \\
    $d_{\bm{o},\bm{p}_{i}^{\ast}}$ & distance from $\bm{o}$ to $\bm{p}^{\ast}_{i}$ \\
    $r^{l}$ & radius of a link \\
    $r^{\mathcal{O}_{i}}$ &
    radius of a spherical object $\mathcal{O}_{i}$ \\
    $\mu$ & coefficient of friction \\
    $\bm{g}$ & unit vector indicating the gravitational direction \\
    $\theta^{i}_{j}$ & joint angle of the $j$th joint on the $i$th finger \\
    $\bm{\theta}^{i}$ & vector of joint angles from the $i$th finger \\
    $\bm{\theta}^{i}_{actv}$ & elements in $\bm{\theta}^{i}$ that parameterize the contact \\
    $\mathbf{Q}_{i}$ & set of variables that parameterize $\{\mathbb{C}_{i}\}$ \\
    $q^{i}$ & a generalized element of $\mathbf{Q}_{i}$ \\
    $\mathbf{Q}^{o}$ & set of variables that parameterize $\{\mathbb{O}\}$ \\ 
    $\kappa$ & kinematic efficiency metric \\
\bottomrule
\end{tabularx}
\end{table}

\subsection{Modelling of hands}
We use a 20-DoF anthropomorphic human right-hand model to explain our proposed method in detail.
Moreover, we also validate our algorithms on a 16-DoF Allegro robotic left-hand model by generating a variety of grasps of everyday objects, which have been tested on a real Allegro robotic hand.

\subsubsection{Human hand model}
The human right-hand model consists of five fingers and a palm.
The geometric shape of finger phalanges is modeled using cylinders, and the palm is modeled by a cuboid (see Fig.~\ref{fig:hand_model}(a)).
The maximal finger length is $92$mm, hand width is $58$mm, and hand length is $120$mm.
The thumb has four joints, including the trapezio\allowbreak metacarpal junction (TM, 2 joints), metacarpo\allowbreak phalangeal junction (MCP, 1 joint), and inter\allowbreak phalangeal (IP, 1 joint).
Each one of the fingers has four joints, corresponding to the rotational joints of metacarpo\allowbreak phalangeal junction (MCP, 2 joints, orthogonal), proximal inter\allowbreak phalangeal junction (PIP, 1 joint), and distal inter\allowbreak phalangeal (DIP, 1 joint).
The palm connects the bases of all fingers and does not have any degrees of freedom.

We denote each of the hand's articulation (degree of freedom) as a \emph{joint} and each segment of the hand (not only finger phalanges but also the palm) as a \emph{link}.
We number fingers sequentially, from the thumb (F1) to the little finger (F5).
Each finger has four phalanges (links) in the anatomical structure.
From the wrist to each fingertip, the first finger link (metacarpal phalanx, L1) belongs to the palm, thus is not considered an independent link.
The proximal, intermediate, and distal phalanges of each finger are independent links, denoted as L2, L3, and L4, respectively.
The palm is also considered as one independent link, represented as ``PALM'' (or L0).
Any phalanx in the model can be retrieved by combining the finger name and link name.
For example, F2L3 refers to the intermediate phalanx (L3) of the index finger (F2).

\subsubsection{Allegro hand model}
\begin{figure}
    \centering
    \includegraphics[width=\columnwidth]{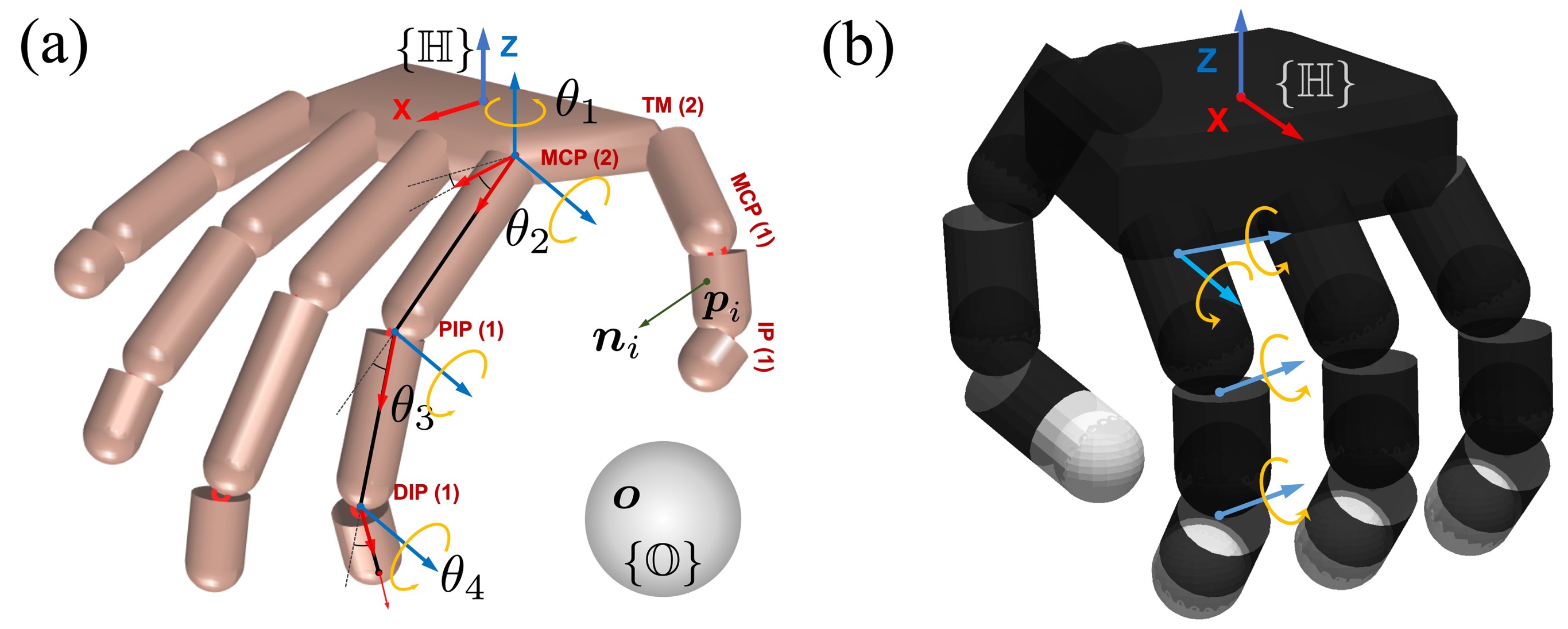}
    \caption{Kinematic models of hands (both models are adapted from \cite{malvezzi2013syngrasp}).
    (a) A 20-DoF human right-hand model.
    (b) A 16-DoF Allegro robotic left-hand model.
    For both models, the hand reference frame $\{\mathbb{H}\}$ is defined at the geometric center of the palm.
    The object reference frame $\{\mathbb{O}\}$ locates at the geometric center of the object.
    Joint reference frames of other fingers are defined following the same convention as the index finger.}
    \label{fig:hand_model}
\end{figure}

The Allegro left-hand model has four fingers and a palm (see Fig.~\ref{fig:hand_model}(b)).
Each finger has one ab-/adduction DoF and three extension/flexion DoFs, defined in the same order as the human hand.
We number fingers sequentially, from the thumb (F1) to the ring finger (F4).
The geometric shape of finger phalanges is also approximated by cylinders, but differs in size from the human hand model.
The finger length is $136.1$mm, hand width is $139.5$mm, and hand length is $247.7$mm.
The hand width is measured as the combined width of all fingers (exclude the thumb) close together (i.e., all fingers are fully extended and the ab-/adduction DoF of all fingers are in $0$ position). The hand width equals the width of the palm for both hand models.

\subsubsection{Definition of reference frame}
Grasp is formulated in the hand reference frame $\{\mathbb{H}\}$, whose origin is located at the geometric center of the palm for both hand models.
Each joint's rotation corresponds to the Z axes of each joint reference frame and is denoted by blue arrow lines (see Fig.~\ref{fig:hand_model}).
Axial directions of phalanges correspond to the X-axis of joint reference frames.
The Y-axes are determined by following the right-hand convention accordingly.

\begin{figure*}[htp!]
    \centering
    \includegraphics[width=\textwidth]{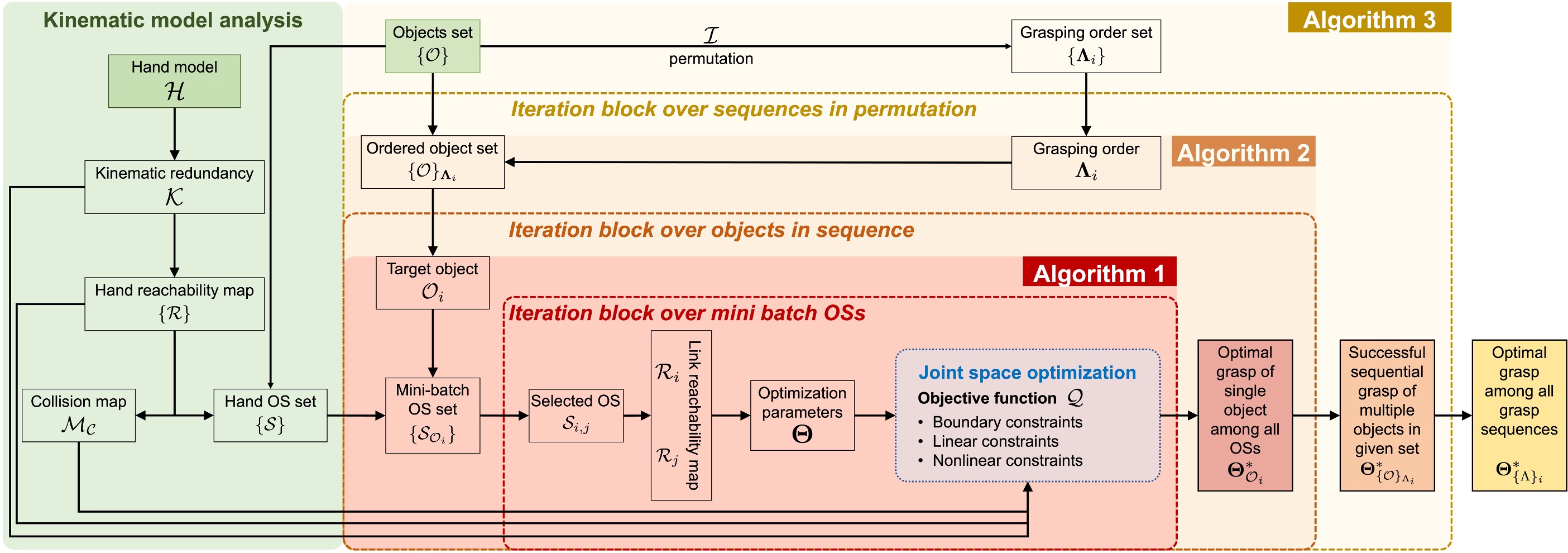}
    \caption{Our proposed framework. It consists of three algorithms: (1) a human-like dexterous grasping algorithm using arbitrary surface regions of the hand (Algorithm~\ref{alg:single_grasp}), (2) an iterative approach for sequential grasping of multiple objects (Algorithm~\ref{alg:sequential_grasp}), and (3) a greedy grasp algorithm to exploit the kinematic redundancy in a sequential grasp (Algorithm~\ref{alg:greedy_grasp}).}
    \label{fig:framework}
\end{figure*}

\subsection{Modelling of objects}
We use elementary geometric shapes, such as spheres and cylinders to approximate the geometric shape of experimental objects \cite{hubbard1996approximating}.
The 6-dimensional pose of the object is described by the \emph{object reference frame} $\{\mathbb{O}\}$, whose origin $\bm{o}$ is located at the geometric center of the object.
For a cylindrical object, we define the Z axis of $\{\mathbb{O}\}$ along the object's central axis. The X and the Y axes are assigned arbitrarily in the orthogonal plane.
We represent the orientation of a cylindrical object using a quaternion $\bm{q}^{\mathcal{O}}=(q_x,q_y,q_z,q_w)$, $\lVert\bm{q}^{\mathcal{O}}\rVert = 1$.
Similarly, we approximate the geometric shape of a complex-shaped object using a group of spheres or cylinders that are fixed in $\{\mathbb{O}\}$.
For example, we approximate a pear (Fig.~\ref{fig:objects}, $\mathcal{O}_{15}$) using two spheres (radius $30$mm, $17$mm) with centers $52$mm apart; and an irregular-shaped bottle using three connected cylinders (Fig.~\ref{fig:objects}, $\mathcal{O}_{16}$).
In case the object is a composition of multiple geometric shapes, we consider the center of the geometric shape to be in contact with the fingers as the center of the whole object, and establish the coordinate system accordingly.
As we focus on generating dexterous grasps, we assume that the target object is light enough for the hand to generate sufficient forces to support the object in a stable grasp, e.g., when the force closure property is satisfied.
Moreover, we assume that the geometric center of the object and its center of mass coincide as the same point $\bm{o}$.

\subsection{Modelling of contact}
Our algorithm allows the contact point $\bm{p}_{i}$ to be assigned to an arbitrary position on the surface of the $i$th link ($\mathcal{L}_{i}$).
A \emph{contact reference frame} $\{\mathbb{C}_{i}\}$ is defined at each contact point.
The contact normal $\bm{n}_{i}$ corresponds to the Z-axis of $\{\mathbb{C}_{i}\}$, is perpendicular to the finger surface curvature of the local contact area.
We parameterize a contact point $\bm{p}\in\mathbf{R}^{3}$ on a given finger phalanx by a tuple of cylindrical coordinates $\rho$, $\phi$, and $\alpha$ in the link reference frame $\{\mathbb{L}_{i}\}$ (Appendix~\ref{app:contact_model}).
The palm is modeled as a cuboid and its inner surface region is modeled as a rectangular-shaped surface.
Contact points on the inner surface of the palm are parameterized by the Cartesian coordinates of $\{\mathbb{C}_{i}\}$.
Our grasp synthesis algorithm initially considers two contact points, which is the minimum number of contacts needed to span an opposition space \cite{arbib1985coordinated,iberall1986opposition}.
Grasping using two opposing hand regions is common for humans \cite{iberall1997human} and suffices most grasping tasks.
Once a grasp is successfully generated, the hand can close its free links to establish more contacts with the grasped object.
These extra contacts could be used to either improve grasping quality or to perform the desired manipulation.

We use the point-contact model with friction and approximate the friction cone by a polyhedral convex cone (Appendix~\ref{app:friction_cone}).
We ignore the torsional moments at the contact, as we are primarily interested in ensuring that the grasped object does not slip off.
Using two contact points to achieve force closure stability condition (Sec.~\ref{sec:force_closure}) prevents translational slip on the contact region.

\section{Proposed Framework}
Our proposed framework (Fig.~\ref{fig:framework}) consists of three parts:
(a) a human-like dexterous grasping algorithm using arbitrary surface regions of the hand (Algorithm~\ref{alg:single_grasp}),
(b) an iterative process for sequential grasping of multiple objects (Algorithm~\ref{alg:sequential_grasp}), and
(c) a greedy-grasping algorithm that aims at maximally exploiting the kinematic redundancy in the task of grasping multiple objects (Algorithm~\ref{alg:greedy_grasp}).

\begin{figure}
    \centering
    \includegraphics[width=\columnwidth]{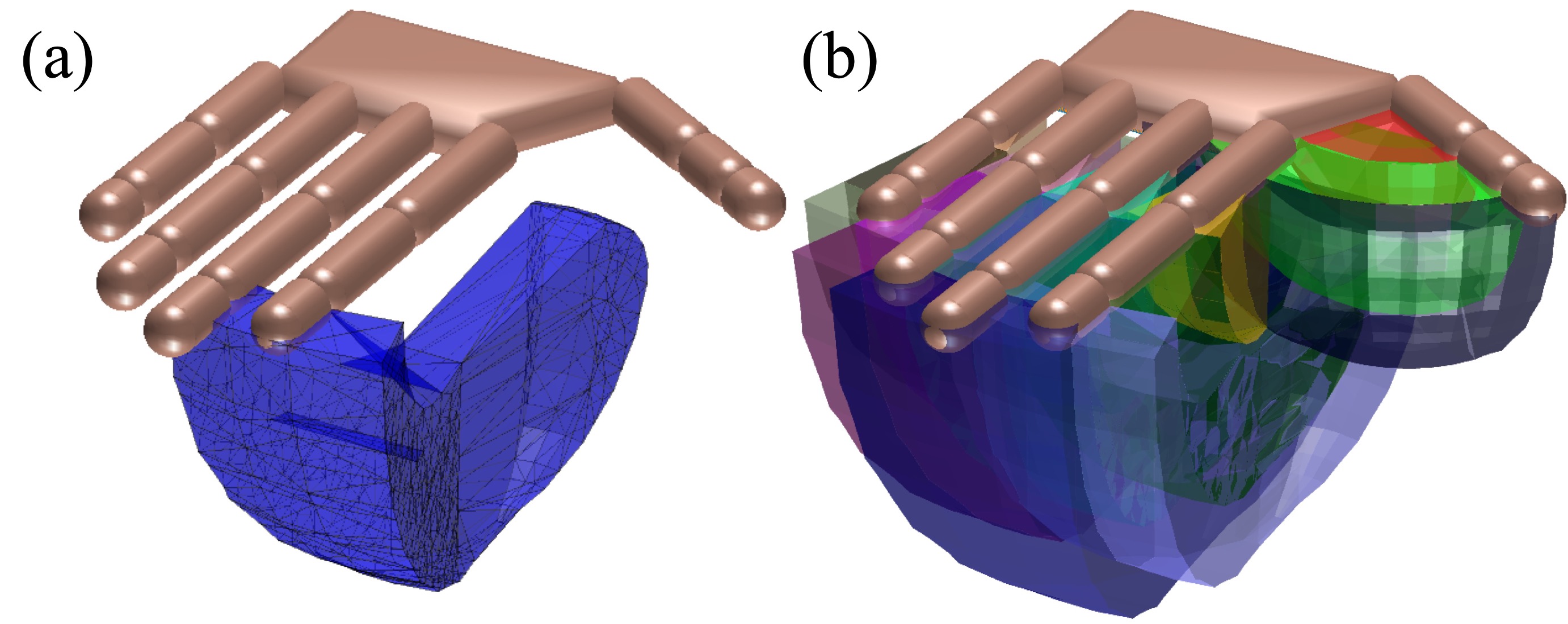}
    \caption{The reachability map sampled for the human hand model.
    (a) The reachable space $\mathcal{R}$ of the index finger's distal phalanx (F2L4).
    (b) The complete reachability map $\{\mathcal{R}\}$ of the hand model.}
    \label{fig:reachability_map}
\end{figure}

\section{Human-like Dexterous Grasp Synthesis}
In this section, we introduce our human-like dexterous grasping algorithm that uses arbitrary hand surface regions.
These grasps are not limited to using fingertips and the palm and can be extended to the complete surface of the hand.

\subsection{Reachability map}\label{sec:reachability_map}
The reachability map is a hand-centered representation of the hand's motion feasibility in Cartesian space.
To construct the hand's reachability map, we uniformly sample each joint of the hand within its motion range and register the corresponding spatial position of each joint calculated by forward kinematics.
Notice that we discriminate the reachable space of one hand link by referring to it as the \emph{reachable space}, and the reachable spaces of the entire hand as the \emph{reachability map}. It is a set of the reachable spaces of all links.
We denote the reachable space of the $i$th joint as $\mathrm{S}_{i}$, which is a continuous three-dimensional surface.
For each finger phalanx link $\mathcal{L}_{i}$, we compute the link's reachable spatial space, $\mathcal{R}_{i}$, composed of the space enclosed by the reachable sets of joints at its both ends, i.e., $\mathrm{S}_{i}$ and $\mathrm{S}_{i+1}$.
The spatial volume bounded in-between is reachable by at least one point on the link, due to the convexity of the link's geometry (a finger phalanx is approximated as a cylinder) and the continuity of the motion range of each DoF.

The link's reachable space $\mathcal{R}_{i}$ is represented as a three-dimensional alpha-shape \cite{edelsbrunner1994three} built up by the sampled Cartesian space points from $\mathrm{S}_{i}$ and $\mathrm{S}_{i+1}$.
As an example, Fig.~\ref{fig:reachability_map}(a) shows the reachable space of the distal phalanx of the index finger (F2L4).
Since the palm is fixed in $\{\mathbb{H}\}$ and has no DoF, its reachable space is represented as the entire rectangular-shaped palm's inner surface.
Fig.~\ref{fig:reachability_map}(b) demonstrates the complete reachability map of the hand, $\{\mathcal{R}\}$.

\subsection{Opposition space}\label{sec:opposition_space}
\begin{figure}[htp!]
\centering
    \includegraphics[width=\columnwidth]{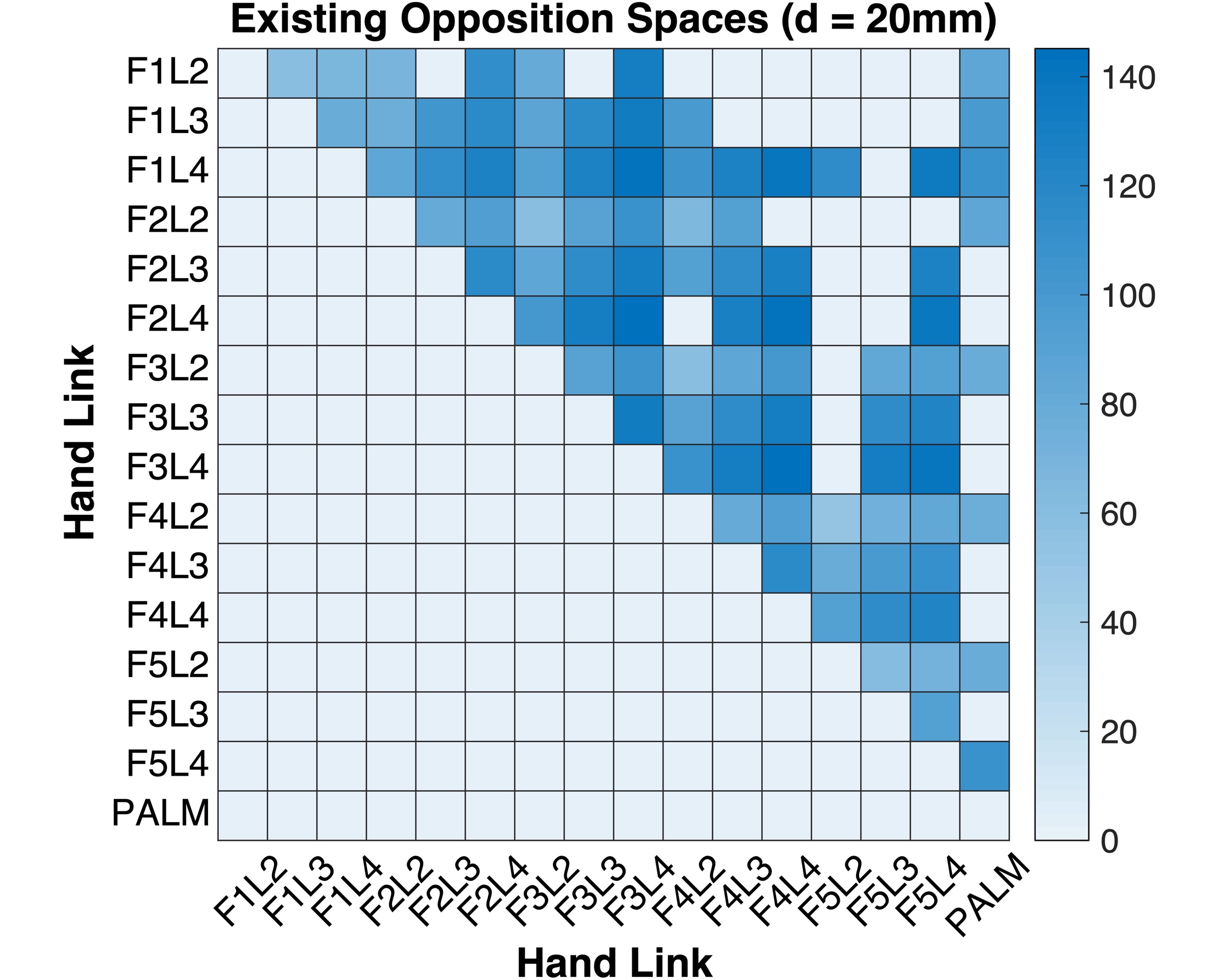}
    \caption{Overview of the geometrically permissive opposition spaces structured inside the human hand model for a grasp distance $d=20$mm ($\mu=0.5$).
    Only the upper triangular of the map is visualized due to symmetry.
    Hand links listed on the label are subject to our naming convention.
    The heatmap of entry values illustrates the largest feasible grasp distance within the corresponding opposition space (all values displayed here are larger than the given desired grasp distance ($d\ge20$mm)).
    Empty entries correspond to unfeasible (e.g., entries on the diagonal) or non-geometrically permissive opposition spaces.}
    \label{fig:opposition_space}
\end{figure}

Given the reachability map of the hand $\{\mathcal{R}\}$, the complete set of opposition spaces spanned by the hand model can be obtained by analyzing the geometric relationship of reachability maps for each pair of links.
We define a generalized opposition space (OS) as the convex hull spanned by a group of reachability maps:
\begin{equation}
    \mathcal{S}=\text{ConvexHull}(\bigcup_{i}\mathcal{R}_{i}).
\end{equation}
Within such an OS, opposing forces can be applied at a certain distance, the value which depends on the spatial relationship of these reachability maps.
Each reachability map represents the complete reachable space of one contact point; thus, to afford a grasp with $N$ contacts, at least $N$ reachability maps are required.
Since our proposed algorithm aims at generating grasps consisting of two contacts, OSes discussed in this paper are spanned by a pair of two reachable spaces $\mathcal{R}_{i}$ and $\mathcal{R}_{j}$, $i\neq j$, denoted by $\mathcal{S}_{i,j}\coloneqq\mathcal{R}_{i}\cup\mathcal{R}_{j}$.
The contact points associated with $\mathcal{S}_{i,j}$ are $\bm{p}_{i}$ and $\bm{p}_{j}$, located on the general links $\mathcal{L}_{i}$ and $\mathcal{L}_{j}$, respectively.
In the following, we drop the subscripts $i$ and $j$ for simplicity and denote the OS candidate as $\mathcal{S}$.
Given a pair of spatial points, $\bm{r}_{i}\in\mathbf{R}^{3}$ from $\mathcal{R}_{i}$ and $\bm{r}_{j}\in\mathbf{R}^{3}$ from $\mathcal{R}_{j}$, we define respectively the minimum and maximum capacity of $\mathcal{S}$:
\begin{equation}\label{eq:os_capacity}
\begin{split}
    & \underline{{C}}_{\mathcal{S}}=\min\|\bm{r}_{i}-\bm{r}_{j}\|,~\overline{{C}}_{\mathcal{S}}=\max\|\bm{r}_{i}-\bm{r}_{j}\|,\\
    & \bm{r}_{i}\in\mathcal{R}_{i},~\bm{r}_{j}\in \mathcal{R}_{j},~i\neq j.
\end{split}
\end{equation}
The boundaries of OS capacity constrain the feasible \emph{grasp distance} $d$, which is the distance between two contact points on the surface of the object $\mathcal{O}$ being grasped.
For example, for a spherical (or cylindrical) object with radius $r$, the minimum grasp distance enabling a force-closure grasp (see Sec.~\ref{sec:force_closure}) is $d = \bm{d}_{\min}(\bm{p}_{i},\bm{p}_{j}) = 2 r \cos(\arctan(\mu))$.
Given a desired $d$, we say the OS ($\mathcal{S}_{d}$) is \emph{geometrically permissive} if:
\begin{equation}\label{eq:opposition_space}
\begin{split}
    & \underline{{C}}_{\mathcal{S}_{d}} \leq d \leq \overline{{C}}_{\mathcal{S}_{d}},\\
    & d \coloneqq \bm{d}_{\min}(\bm{p}_{i},\bm{p}_{j}),~\bm{p}_{i}\in\mathcal{R}_{i},~\bm{p}_{j}\in\mathcal{R}_{j},~i\neq j.
\end{split}
\end{equation}
For example, Fig.~\ref{fig:opposition_space} lists all feasible OSes constructed by the human hand model for a grasp distance $d=20$mm.
$\mathcal{S}_{d}$ can then be used to synthesize a grasp for a target object $\mathcal{O}$ at a grasp distance $d$.
We use $\{\mathcal{S}_{d}\}$ to represent the complete set of all geometrically permissive OSes for $d$.
Notice that such property is related to the object model $\mathcal{O}$ and being geometrically permissive is a sufficient but not necessary condition for an OS to enable a feasible grasp.
In the following, we simply denote $\mathcal{S}_{d}$ as $\mathcal{S}$, since all OSes must be geometrically permissive to afford a feasible grasp.

\subsection{Grasp synthesis as constrained optimization}
We formulate the grasp synthesis problem as a constrained optimization problem.
As soon as an OS candidate $\mathcal{S}$ is selected, the links used for grasping are determined.
We denote the set of all kinematic variables that parametrize the local contact frame $\{\mathbb{C}_{i}\}$ as $\mathbf{Q}_{i}$.
Such elements include (1) joint angles that precede $\bm{p}_{i}$ along the kinematic chain and (2) the cylindrical coordinates that model $\bm{p}_{i}$ in the local reference frame (see Appendix~\ref{app:contact_model}).
An element of $\mathbf{Q}_{i}$ is represented as $q$.
We consider three types of constraints: boundary constraints, collision constraints (inequality constraints), and contact constraints (equality constraints).

\subsubsection{Boundary constraints}\label{sec:boundary_constraints}
\paragraph*{Kinematic variables} the generalized parameters $q$ in $\mathbf{Q}_{i}$ parameterizes the contact $\bm{p}_{i}$. It could be a joint angle or a cylindrical coordinate, both subject to boundary constraints.
\begin{equation}\label{eq:constraint_kinematic_boundary}
q\in [\underline{q},\overline{q}],~\forall q\in \mathbf{Q}_{i},~i=1,\dots,{N}_{c},
\end{equation}
with $\underline{q}$ and $\overline{q}$ being the corresponding lower and upper bound of $q$.
${N}_{c}$ is the total number of contacts.

\paragraph*{Object center} a target object is described in $\{\mathbb{H}\}$, and it can only move inside the OS to guarantee the existence of contacts.
An accurate closed-form description of such a bounded region is often difficult to obtain, as the opposition space is an irregular spatial volume in general, and the object's geometry also affects the boundary of this region.
We approximate this constraint by restricting the object center $\bm{o}\in\mathbf{R}^{3}$ inside the entire opposition space:
\begin{equation}\label{eq:constraint_object_center}
    \bm{o}\in \mathcal{S}.
\end{equation}
An approximation of the geometric shape of $\mathcal{S}$ can be obtained by using the circumscribed spatial cuboid region of $\mathcal{S}$ to simplify the calculation.

\subsubsection{Collision constraints} \label{sec:collision_constraints}
\begin{figure}[htp!]
\centering
    \includegraphics[width=\columnwidth]{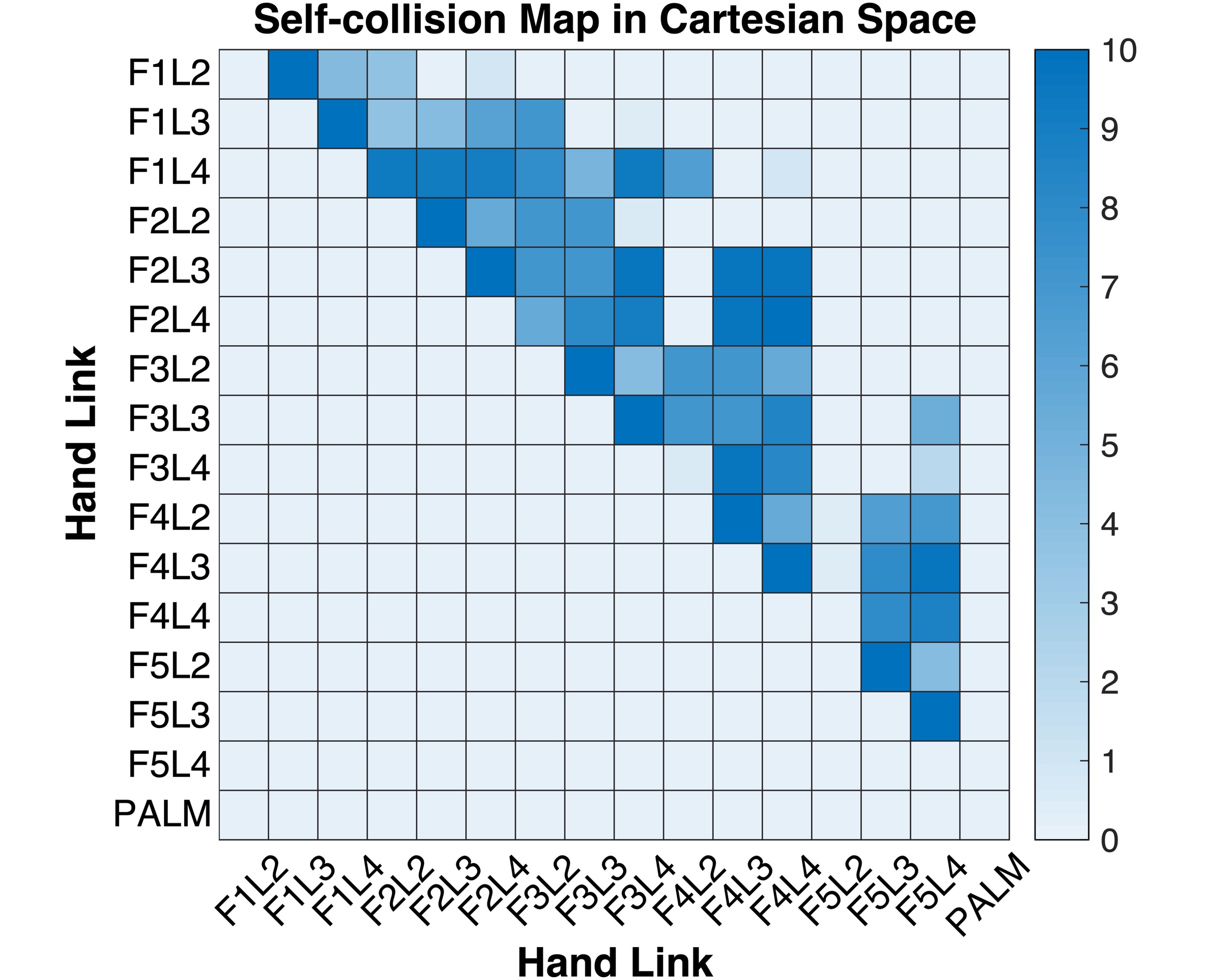}
    \caption{Self-collision map of the human hand model, visualized as a heatmap.
    The radius of all finger phalanges is $r^{l}=5$mm.
    Hand links listed on the label are subject to our naming convention.
    The entry value illustrates the maximum potential collision depth in Cartesian space between two links.
    Empty entries indicate collision-free link pairs.}
    \label{fig:collision_map}
\end{figure}

A feasible grasp must be collision-free.
We consider three types of collision: (1) link-object collision, (2) link-link collision, and (3) object-object collision.

\paragraph*{Link-object collision} to avoid collision between a target object and hand links, the spatial distance between the object center $\bm{o}$ and the center of an arbitrary link $\mathcal{L}_{i}$ must satisfy:
\begin{equation}\label{eq:constraint_link_object_collision}
    \bm{d}(\mathcal{O},\mathcal{L}_{i}) \geq r_{l} + d_{\bm{o},\bm{p}_{i}^{\ast}},~ i=1,\dots,{N}_{l}.
\end{equation}
$r_{l}$ is the distance from the link center (or central axis) to the contact on the link surface, $d_{\bm{o},\bm{p}_{i}^{\ast}}$ is the distance from the object center $\bm{o}$ to $\bm{p}_{i}^{\ast}$, i.e., the projection of $\bm{p}_{i}$ onto the object surface along $\overrightarrow{\bm{p}_{i} \bm{o}}$.
The distance $\bm{d}(\mathcal{O},\mathcal{L}_{i})$ can be calculated using a sampling-based approach by creating $N$ uniformly sampled points on the central axis to represent a cylindrical shape \cite{el2013generation}.
If $\mathcal{L}_{i}$ is a finger phalanx approximated by $N_\mathcal{L}$ samples, $\bm{d}(\mathcal{O},\mathcal{L}_{i})$ can be calculated as $N_\mathcal{L}$ distance from the center $\bm{o}$ of a spherical object to each sample point.
$r_{l}$ is the radius of the phalanx cylinder, and $d_{\bm{o},\bm{p}_{i}^{\ast}}$ is the radius of the object sphere.
For a cylindrical object approximated by $N_\mathcal{O}$ samples, $\bm{d}(\mathcal{O},\mathcal{L}_{i})$ has a total number of $N_\mathcal{L}\times N_\mathcal{O}$ distances, and $d_{\bm{o},\bm{p}_{i}^{\ast}}$ is the radius of the cylinder.
If $\mathcal{L}_{i}$ represents the palm, $r_{l}$ is half the thickness of the palm (a cuboid), and $\bm{d}(\mathcal{O},\mathcal{L}_{i})$ is calculated as the projection distance of $\bm{o}$ (for a cylindrical object, it is each one of the samples on its central axis) onto the palm plane, along the palm's surface normal direction.
\paragraph*{Link-link collision} constraints of this type ensure that the hand is in a self-collision-free configuration.
The overlapping regions of the hand reachability map (see Fig.~\ref{fig:reachability_map}(b)) indicate potential collision among links in Cartesian space.
A pair of links $\mathcal{L}_{i}$ and $\mathcal{L}_{j}$ may potentially collide if the minimum spatial distance between their corresponding reachable spaces $\mathcal{R}_{i}$ and $\mathcal{R}_{j}$ is smaller than $r^{l}_{i,j}=r^{l}_{i}+r^{l}_{j}$.
The distance $r^{l}_{i}$ or $r^{l}_{j}$ is the cylinder radius for a finger phalanx link or half the thickness of the cuboid for the palm.

We construct a \emph{self-collision map} $\mathcal{M}_\mathcal{C}$ to register all potential collisions between links (including palm) in the hand model, by analyzing the reachability map set of the hand $\{\mathcal{R}\}$.
Collisions that exist for all pairs of links $(\mathcal{L}_{i}$, $\mathcal{L}_{j})$ are checked using sampling based approach to construct $\mathcal{M}_\mathcal{C}$.
For a pair of finger phalanx links, each link has ${N}_{k}$ uniformly and orderly sampled points from the head to the tail of the link on its central axis, $\{c^{i}_{k}\}$ and $\{c^{j}_{k}\}$, $k=1,2,\dots,{N}_{k}$, respectively.
Then the constraint is formulated as a vector of ${N}_{k}$ pairwise distances between sampled points and must satisfy:
\begin{equation}\label{eq:constraint_link_link_collision}
\begin{split}
    & \bm{d}(\mathcal{L}_{i},\mathcal{L}_{j}) \coloneqq [d_{c^{i}_{1},c^{j}_{1}},\dots,d_{c^{i}_{k},c^{j}_{k}},\dots,d_{c^{i}_{{N}_{k}},c^{j}_{{N}_{k}}}]^{\intercal}\succeq 
    {\bm{r}^{l}_{i,j}},\\
    & \bm{r}^{l}_{i,j} \in \mathbf{R}^{{N}_{k}\times 1},~k=1,2,\dots,{N}_{k},\\
    & i=1,2,\dots,{N}_{l},~j=1,2,\dots,{N}_{l},~i \neq j.\\
\end{split}
\end{equation}
For a pair of finger phalanges, each entry in $\bm{r}^{l}_{i,j}$ is calculated as ${r}^{l}_{i,j} = r^{l}_{i}+r^{l}_{j}$.
The collision between a finger phalanx link $\mathcal{L}_{i}$ and the palm is constrained by forcing the projection distance of each sample $c^{i}_{k}$ to the palm surface larger than $\bm{r}^{l}_{i,j}=r^{l}_{i}$.
The self-collision map can be visualized as a heatmap.
For example, Fig.~\ref{fig:collision_map} shows the self-collision map of the human hand model.
The entry values of the map represent the maximum collision depth, i.e., $\max(r^{l}_{i,j}-\bm{d}(\bm{r}_{i},\bm{r}_{j}))$, $\bm{r}_{i}\in\mathcal{R}_{i}$, $\bm{r}_{j}\in \mathcal{R}_{j}$, $i\neq j$.
Only positive values indicate a potential collision.
The upper bound of this depth is ${r}^{l}_{i,j}$.

\paragraph*{Object-object collision}
when grasping multiple objects (see Sec.~\ref{sec:multiple_objects_grasping}), any two objects $\mathcal{O}_{i}$ and $\mathcal{O}_{j}$ (can be grasped objects or target objects) should not collide:
\begin{equation}\label{eq:constraint_object_object_collision}
    \begin{split}
    & \bm{d}(\mathcal{O}_{i},\mathcal{O}_{j}) \geq d_{\bm{o}_{i},\bm{o}_{j}^{\ast}} + d_{\bm{o}_{j},\bm{o}_{i}^{\ast}},\\
    & ~i=1,2,\dots,{N}_{o},~j=1,2,\dots,{N}_{o},~i \neq j,
    \end{split}
\end{equation}
where $\bm{o}_{j}^{\ast}$ is the projection of $\bm{o}_{j}$ onto the surface of $\mathcal{O}_{i}$ along the line of $\overrightarrow{\bm{o}_{i}\bm{o}_{j}}$, and $d_{\bm{o}_{i},\bm{o}_{j}^{\ast}}$ is the distance from object center $\bm{o}_{i}$ to this projected point.
For a spherical object, $d_{\bm{o}_{i},\bm{o}_{j}^{\ast}}$ equals the radius $r^{\mathcal{O}^{i}}$.
A cylindrical object can be represented by a sequence of $N_\mathcal{O}$ sample points on its central axis, as introduced previously.
In this case, for each one of the $N_\mathcal{O}$ samples, $d_{\bm{o}_{i},\bm{o}_{j}^{\ast}}$ is the radius of the cylinder.
For irregular-shaped objects, $d_{\bm{o}_{i},\bm{o}_{j}^{\ast}}$ can be calculated if an explicit surface model is available.
Alternatively, the distance can also be estimated based on an implicit representation of the object surface \cite{el2015computing}.
The sampling-based approach could also be applied to calculate the distance for irregular-shaped objects.
For instance, approximate the shape of each object by sampling multiple points on the surface.
The minimum distance between samples on the surface of $\mathcal{O}_{i}$ and samples on $\mathcal{O}_{j}$ can be used as an estimation of $\bm{d}(\mathcal{O}_{i},\mathcal{O}_{j})$.
In this work, we calculate $\bm{d}(\mathcal{O}_{i},\mathcal{O}_{j})=r^{\mathcal{O}_{i}}+r^{\mathcal{O}_{j}}$ for a pair of spherical objects.
Give a cylindrical object and a spherical object, the distance becomes a vector of length $N_\mathcal{O}$ with each entry representing the Euclidean distance between the spherical object center and one of the $N_\mathcal{O}$ sample points on the central axis of the cylindrical object.
Similarly, for a pair of cylindrical objects, the distance vector has $N_{\mathcal{O}_{i}}\times N_{\mathcal{O}_{j}}$ entries.

\subsubsection{Contact constraints}\label{sec:contact_constraints}
Contact constraints (1) ensure contacts are established between the hand link and the target object, and (2) provide a grasping stability guarantee.

\paragraph*{In-contact constraint} this equality constraint guarantees that each contact point $\bm{p}_{i}$ on the surface of the hand link $\mathcal{L}_{i}$ must also present on the object surface:
\begin{equation}\label{eq:constraint_in_contact}
    \bm{d}(\mathcal{O},\bm{p}_{i}) = d_{\bm{o},\bm{p}^{\ast}_{i}},~i=1,2,\dots,{N}_{c}.
\end{equation}
For a spherical object, $\bm{d}(\mathcal{O},\bm{p}_{i}) \coloneqq d_{\bm{o},\bm{p}_{i}}$, and $\bm{p}^{\ast}_{i}$ is the projection of $\bm{p}_{i}$ onto the sphere surface along the vector $\overrightarrow{\bm{p}_{i} \bm{o}}$.
$d_{\bm{o},\bm{p}^{\ast}_{i}}$ is the radius of the object.
For a cylindrical object, $\bm{d}(\mathcal{O},\bm{p}_{i})$ is the distance from $\bm{p}_{i}$ to the line segment of the cylinder central axis, and $\bm{p}^{\ast}_{i}$ is the projection of $\bm{p}_{i}$ onto the cylinder surface along the cylinder radical direction.

\paragraph*{Stability constraint}\label{sec:force_closure}
we use force closure property to ensure grasp stability.
It states that any external wrenches applied on the grasped object can be counterbalanced by a resultant wrench applied through contacts.
In a point contact model with friction, contact force must lie inside the corresponding friction cone, such that Coulomb's law is satisfied and slip does not occur on the contact point \cite{kraus1998analysis}.
The force $\bm{f}^{i}$ generated by the $i$th contact point and lying inside its friction cone can be approximated as a linear combination of all edges of the ${N}_{f}$-sided friction cone, centered around the contact normal $\bm{n}_{i}$ \cite{miller1999examples}:
\begin{equation*}
\bm{f}^{i} \approx \sum_{n=1}^{{N}_{f}} \lambda^{i}_{n} \mathbf{f}^{i}_{n},~\sum_{n=1}^{{N}_{f}} \lambda^{i}_{n} = 1,~\lambda^{i}_{n}\geq 0,~i=1,\dots,N_c,~n=1,\dots,N_f.
\end{equation*}
The calculation related to friction cone approximation is explained in Appendix~\ref{app:friction_cone}.
The torque generated by the force component $\mathbf{f}^{i}_{n}$ on the object center is
$\bm{\tau}^{i}_{n} = \bm{d}^{i} \times \mathbf{f}^{i}_{n}$,
where $\bm{d}^{i}$ is the vector from the object center $\bm{o}$ to the $i$th contact point $\bm{p}_{i}$, and the associated primitive contact wrench:
$\bm{w}^{i}_{n} = %
    \begin{pmatrix}
        \mathbf{f}^{i}_{n}\\ 
        \bm{\tau}^{i}_{n}
    \end{pmatrix}$.
A grasp has force closure if and only if the origin of the wrench space lies inside the convex hull constructed by the contact wrenches \cite{montana1991condition}.
This can be formulated as a convex optimization problem by determining a set of real positive coefficients $\{\varphi^{i}_{n}\}$ that satisfy:
\begin{equation}\label{eq:constraint_force_closure}
    \begin{split}
        & \exists~\varphi^{i}_{n}\in \mathbf{R},~\varphi^{i}_{n}\geq 0,~\sum_{i,n}\varphi^{i}_{n}=1,\\
        & i=1,2,\dots,{N}_{c},~n=1,2,\dots,{N}_{f},\\
        & \text{s.t.}~\sum_{i,n}\varphi^{i}_{n}\bm{w}^{i}_{n}=0.
    \end{split}
\end{equation}
The coefficients $\{\varphi^{i}_{n}\}$ are integrated into the grasp synthesis optimization as optimization variables.

\paragraph*{Other constraints} the quaternion parameterizing the object orientation must be constrained such that:
\begin{equation}\label{eq:quaternion_norm}
    \lVert\bm{q}^{\mathcal{O}}\rVert = 1.
\end{equation}

\subsubsection{Objective function}\label{sec:objective_function}
We formulate the objective function by considering mainly the grasp quality cost and the movement economy cost.

\paragraph*{Grasp quality cost} although satisfying the force-closure constraint guarantees grasping stability, the force-closure property could be of poor quality and is easily violated \cite{zhu2004planning}.
This margin of violating force closure property is geometrically equivalent to the minimum distance from the surface of the convex hull to the origin in the grasp wrench space \cite{borst2004grasp}, and a handful of metrics have been formulated for grasp quality evaluation
\cite{roa2015grasp,aleotti2010interactive}.
We use two metrics proposed by \cite{klein2020predicting} as costs in both force and torque aspects.
Such metrics have been used in the analysis of human precision grasps of 3D objects and proved to be effective.
In a grasp with two contacts, the quality metric can be calculated by analyzing the relationship between the force application directions \cite{nguyen1988constructing}.
In ideal cases, the contact normal should be aligned with the grasp axis (i.e., the line segment connecting two contact points), such that both applied contact forces lie along the central axes of the corresponding friction cones, and the wrench space has the largest margin under the force closure constraint.
The force cost $C_{f}$ penalizes the deviance between the grasp axis and the contact normal:
\begin{equation}
\begin{split}
    C_{f} = & \arctan(\|\bm{n}_{1}\times(\bm{p}_{2}-\bm{p}_{1})\|,~ \bm{n}_{1}\cdot (\bm{p}_{2}-\bm{p}_{1}))\\
    + & \arctan(\|\bm{n}_{2}\times(\bm{p}_{1}-\bm{p}_{2})\|,~\bm{n}_{2}\cdot (\bm{p}_{1}-\bm{p}_{2})),
\end{split}
\end{equation}
where $\bm{n}_{1}$ and $\bm{n}_{2}$ are the directions of force applications, and $\bm{p}_{1}$ and $\bm{p}_{2}$ are the Cartesian coordinates of contacts represented in $\{\mathbb{H}\}$.
The torque cost $C_{t}$ penalizes the magnitude of the total torque to avoid rotation of the grasped object:
\begin{equation}
    C_{t} = \|(\bm{o}-\bm{p}_{1})\times(-\bm{g})+(\bm{o}-\bm{p}_{2})\times(-\bm{g})\|,
\end{equation}
where $\bm{g}$ is a unit vector in the gravitational direction.

\paragraph*{Movement economy cost} this cost aims to suppress the unnecessary motion of fingers to avoid occupying of adjacent OSes constructed by other links.
$\bm{\theta}^{i}_{actv}$ is a vector consisting of all joint angles affecting the contact point on the $i$th finger, and $\Delta\bm{\theta}^{i}_{actv}$ indicates the change in joint angles compared to the initial position (fully extension or no ab-/adduction).
We add movement economy cost to penalize excessive joint angles:
\begin{equation}
    C_{q} = \sum_{i=1}^{{N}_{c}} {\Delta\bm{\theta}^{i}_{actv}}^\intercal~Q_{i}~ {\Delta\bm{\theta}^{i}_{actv}},~i=1,\dots,{N}_{c},
\end{equation}
where ${N}_{c}$ is the number of contacts, and $Q_{i}$ is a positive definite diagonal matrix that prioritizes the motion of different finger joints.

\paragraph*{Objective function} by integrating the above-mentioned costs, we formulate the objective function as:
\begin{equation}\label{eq:objective_function_single}
    \mathcal{Q} = \lambda \cdot (C_{f}+C_{t}) + (1-\lambda)\cdot C_{q},~\lambda\in [0,1].
\end{equation}
The weighting parameter $\lambda$ balances costs in both aspects.

\subsubsection{Grasp synthesis formulation}\label{sec:optimization_formulation}
Considering the constraints and objective function described above, we formulate our grasp synthesis problem as follows:
\begin{equation}\label{eq:grasp_synthesis}
    \begin{split}
        & \mathbf{\Theta}^{\ast} = \arg\min_{\mathbf{\Theta}}~\mathcal{Q} \\
        & \text{subject to}~
        (\ref{eq:constraint_kinematic_boundary}),~
        (\ref{eq:constraint_object_center}),~
        (\ref{eq:constraint_link_object_collision}),~
        (\ref{eq:constraint_link_link_collision}),~
        (\ref{eq:constraint_object_object_collision}),~
        (\ref{eq:constraint_in_contact}),~
        (\ref{eq:constraint_force_closure}),~
        (\ref{eq:quaternion_norm})
    \end{split}
\end{equation}
The optimization variable $\mathbf{\Theta}$ consists of (1) $\mathbf{Q}^\mathcal{O}$, parameters that parameterize the object reference frame $\{\mathbb{O}\}$, including object center position and quaternion, (2) $\mathbf{Q}_{i}=\{\bm{\theta}^{i}_{actv},\psi_{i},\alpha_{i}\}$, $i=1,\dots,{N}_{c}$, the kinematic variables that parameterize each contact reference frame $\{\mathbb{C}_{i}\}$ on the hand link, and (3) $\{\varphi^{i}_{n}\}$, $i=1,\dots,N_c$, $n=1,\dots,N_f$, the coefficients associated with the edges of friction cones on contact points.

\begin{algorithm}[ht!]
\SetAlgoLined
\KwData{hand model $\mathcal{H}$\;
kinematic redundancy set $\mathcal{K}$\;
target object model $\mathcal{O}$\;}
\KwResult{hand configuration $\mathcal{H}^{\ast}$ and object pose $\mathcal{O}^{\ast}$ in the generated optimal grasp, described by the corresponding optimization variables $\mathbf{\Theta}^{\ast}$\;}
 initialization\;
 \eIf{$\mathcal{K}\neq\emptyset$}
 {
 $\{\mathcal{R}\}$ $\leftarrow$ reachability\_map($\mathcal{H}$,~$\mathcal{K}$)\;
 $\{\mathcal{S}\}$ $\leftarrow$ opposition\_spaces($\{\mathcal{R}\}$)\;
 $\{\mathcal{S}_{\mathcal{O}}\}$ $\leftarrow$ OS\_candidates($\{\mathcal{S}\}$,~$\mathcal{O}$)\;
        $\{\mathbf{\Theta}\}$ = $\emptyset$\;
        k = 1\;
        \While{
        $\{\mathcal{S}_{\mathcal{O}}\}\neq\emptyset$}
        {
        $\mathcal{S}_{i,j}\leftarrow$ select\_OS($\{\mathcal{S}_{\mathcal{O}}\}$)\;
        $\mathbf{Q}_{i}$,~$\mathbf{Q}_{j}$ $\leftarrow$ extract\_parameters($\mathcal{L}_{i},~\mathcal{L}_{j},~\mathcal{S}_{i,j}$)\;
        $\mathbf{\Theta}_{k}\leftarrow$ grasp\_synthesis($\mathbf{Q}^{o}$,$\mathbf{Q}_{i}$,$\mathbf{Q}_{j}$,$\{\varphi^{i}_{n}\}$,$\{\varphi^{j}_{n}\}$)\;
        $\{\mathcal{S}_{\mathcal{O}}\} \leftarrow \{\mathcal{S}_{\mathcal{O}}\} \setminus \mathcal{S}_{i,j}$\;
        k = k + 1\;
            \eIf{$\mathbf{\Theta}_{k} \neq \emptyset$}
            {
            $\{\mathbf{\Theta}\} \leftarrow \{\mathbf{\Theta}\} \bigcup \mathbf{\Theta}_{k}$\;
            }
            {
            continue\;
            }
        }
        $\mathbf{\Theta}^{\ast}$ = minimum\_cost($\{\mathbf{\Theta}\}$)\;
        ($\mathcal{H}^{\ast}$, $\mathcal{O}^{\ast}$, $\mathcal{K}$) = update\_configuration($\mathcal{H}$,~$\mathcal{O}$,~$\mathbf{\Theta}^{\ast}$)\;
        return $\mathcal{H}^{\ast}$, $\mathcal{O}^{\ast}$\;
 }
 {return $\mathcal{H}$,~$\mathcal{O}$;}
 \caption{Human-like dexterous grasp synthesis.}\label{alg:single_grasp}
\end{algorithm}
We summarize our proposed algorithm in Algorithm~\ref{alg:single_grasp}.
Given the hand model $\mathcal{H}$ and the kinematic redundancy set $\mathcal{K}$ that contains all available degrees of freedoms in the model, the first step after initialization is to construct the reachability map set $\{\mathcal{R}\}$ for all hand links.
Then, the set of geometrically permissive opposition spaces $\{\mathcal{S}_{\mathcal{O}}\}$ are determined together with the object model $\mathcal{O}$.
Any OS from this set, $\mathcal{S}$, can be selected as a candidate for grasp synthesis.
Once the candidate $\mathcal{S}$ is selected, the corresponding reachable spaces, $\mathcal{R}_{i}$ and $\mathcal{R}_{j}$, along with the associated hand links $\mathcal{L}_{i}$ and $\mathcal{L}_{j}$ are also determined.
Each link is supposed to establish one contact on the object's surface.
Kinematic parameters of each contact ($\mathbf{Q}_{i}$ and $\mathbf{Q}_{j}$), object pose parameter set $\mathbf{Q}^\mathcal{O}$, as well as the coefficients $\{\varphi^{i}_{n}\}$ and $\{\varphi^{j}_{n}\}$ are added to the optimization variable set.
If multiple opposition space candidates have been selected resulting in more than one feasible solution, the optimal grasp is determined as the one that leads to the minimum objective function value.
Finally, variables in the optimal solution $\mathbf{\Theta}^{\ast}$ are extracted to update the kinematic configuration of both the hand model and the object model to realize the grasp.
After obtaining a successful grasp, the employed DoFs in the grasp are excluded from $\mathcal{K}$.
Such DoFs should not serve as available variables in future tasks.

\section{Multiple Objects Grasping}\label{sec:multiple_objects_grasping}
We start by presenting an algorithm to grasp multiple objects in sequence (Sec.~\ref{subsec:sequential_grasping}).
Then, we propose a kinematic efficiency metric and an associated strategy to facilitate the exploitation of the kinematic redundancy in sequential task planning (Sec.~\ref{subsec:greedy_grasping}).

\subsection{From 1 to N: sequential grasping of multiple objects}\label{subsec:sequential_grasping}
We consider a set of $N_o$ target objects, denoted as $\{\mathcal{O}_{i}\},~i=1,2,\dots,N_o$, and the robotic hand grasps them following the order $\bm{\Lambda}$, which is an ordered list of objects indices.
For example, a total of $N_o=5$ objects can be grasped in the order of $\mathcal{O}_{3}$, $\mathcal{O}_{1}$, $\mathcal{O}_{5}$, $\mathcal{O}_{4}$, $\mathcal{O}_{2}$, given the grasp order $\bm{\Lambda}=[3,1,5,4,2]$.

\subsubsection{The ``Analysis-Grasp" iterative process}
\begin{figure*}[ht]
\centering
\includegraphics[width=\textwidth]{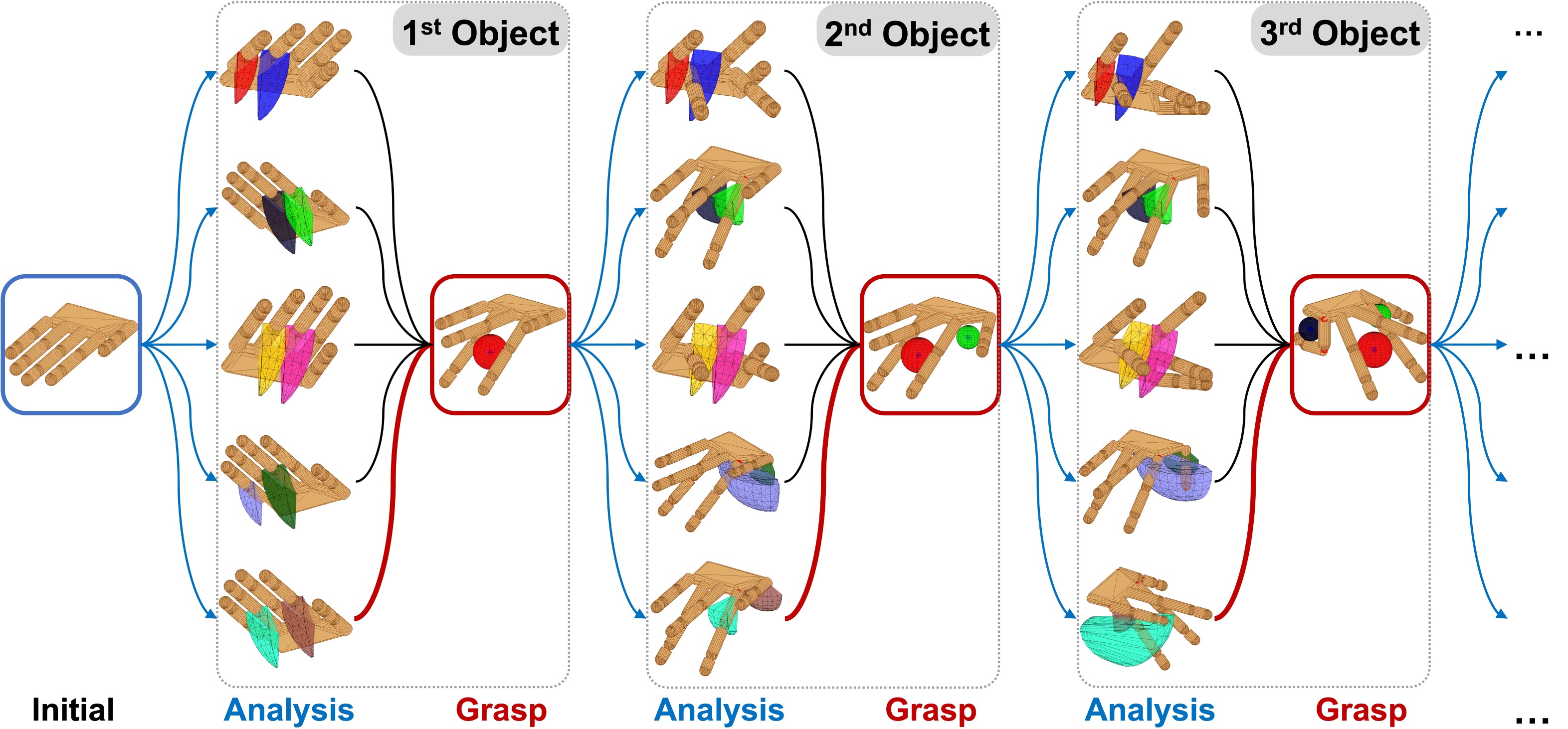}
\caption{The \emph{Analysis-Grasp} iterative process for sequentially grasping multiple objects.
In this example, five OSes are selected as the candidates to grasp each object by analyzing the reachability map updated at each step.
The OS that leads to the optimal solution is selected as the final grasp.}
\label{fig:sequential_grasping}
\end{figure*}

Taking advantage of our proposed grasp synthesis algorithm, we reformulate the problem of grasping multiple objects as a sequence of single-object grasping sub-problems.
Each sub-problem deals with the grasping of one single object, and it consists of two steps.
In the \emph{analysis} step, the robotic hand constructs (if this is the first grasp in the sequence) or updates the hand reachability map $\{\mathcal{R}\}$, and selects one or multiple geometrically permissive opposition spaces $\mathcal{S}$ as candidates for subsequent grasp synthesis.
In the \emph{grasp} step, each candidate $\mathcal{S}$ is used to synthesize a grasp.
The optimal solution is used to update the configuration of the hand and the object, and also the kinematic redundancy set $\mathcal{K}$.
Fig.~\ref{fig:sequential_grasping} illustrates this iterative process by demonstrating a three-object grasping problem with five OS candidates selected in the grasp of each object.
The robotic hand repeats this iterative process, until either all given objects are grasped, or no more opposition spaces can be employed to generate a feasible grasp.

\subsubsection{Selection of opposition spaces}
To determine the optimal grasp for each target object $\mathcal{O}_{i}$, it is possible to test all feasible OS candidates in the set $\{\mathcal{S}_{\mathcal{O}_{i}}\}$ using Algorithm~\ref{alg:single_grasp}.
However, this inevitably demands a large computational cost.
It could also result in an optimal but inefficient grasp, e.g. an OS having a large space capacity has been used to grasp a tiny object.
This may be problematic for the subsequent grasp of potentially large objects.
To alleviate these two potential issues, in each sub-problem, we only select a mini-batch of OS candidates that correspond to the $N$ smallest capacity values $\overline{{C}}_{\mathcal{S}}$ (Eq.~\ref{eq:os_capacity}) among all OSes ($N<|\{\mathcal{S}_{\mathcal{O}_{i}}\}|$).
In case no feasible solutions can be obtained using these $N$ candidates, more feasible OSes can be selected according to the ascending order of their capacity values.

\subsubsection{Update of hand model}
Once the target object has been successfully grasped, the employed OS is excluded from the total OS set $\{\mathcal{S}\}$, and all DoFs affecting the contacted hand links are considered fixed and excluded from $\mathcal{K}$.
Our algorithm enables one link to be reused in multiple grasps, as long as it spans opposition spaces with other links, even though these contacted links are considered to be fixed.
We summarize the algorithm for sequential multiple objects grasping in Algorithm~\ref{alg:sequential_grasp}.
Notice that Algorithm~\ref{alg:sequential_grasp} tests the mini-batch OS candidates and returns an optimal solution.
If no feasible solution is found after testing all OS candidates, the hand then continues to grasp the next object.
Successfully grasped objects are added to the set $\{\mathcal{O}^{\ast}_{i}\}$.
The hand model $\mathcal{H}$ and its kinematic redundancy set $\mathcal{K}$ are updated at the end of each iteration.
Notice that each grasped object potentially brings in novel object-object collision constraints (Eq.~\ref{eq:constraint_object_object_collision}) into subsequent grasps.

\begin{algorithm}
\SetAlgoLined
\KwData{hand model $\mathcal{H}$\;
kinematic redundancy set $\mathcal{K}$\;
set of target object models $\{\mathcal{O}_{i}\}$, $i=1,\dots,N_o$\;
grasping order $\bm{\Lambda}$\;}

\KwResult{optimal grasp configuration, including the configuration of hand $\mathcal{H}^{\ast}$ and the configuration of each grasped object $\{\mathcal{O}^{\ast}_{i}\}$, $i=1,\dots,N_g$\;}

initialization\;
k = 1\;
$\{\mathcal{O}^{\ast}_{i}\} = \emptyset$\;
\While{$\{\mathcal{O}_{i}\}\neq\emptyset$ $\And$ $\mathcal{K}\neq\emptyset$}
{
    $\mathcal{O}_{k}$ $\leftarrow$ select\_object($\{\mathcal{O}_{i}\}$, ${\bm{\Lambda}_{k}}$)\;
    
    $\mathbf{\Theta}_{k}^{\ast}$ $\leftarrow$ Algorithm\_1($\mathcal{H}$, $\mathcal{K}$, $\mathcal{O}_{k}$)\;
    $\{\mathcal{O}_{i}\}$ $\leftarrow$ $\{\mathcal{O}_{i}\}$ $\setminus$ $\mathcal{O}_{k}$\;
    k = k + 1\;
    \If{$\mathbf{\Theta}_{k}^{\ast} \neq \emptyset$}
    {
    ($\mathcal{H}$,~$\mathcal{O}^{\ast}_{k}$) $\leftarrow$ update\_configuration($\mathcal{H}$,~$\mathcal{O}_{k}$,~$\mathbf{\Theta}_{k}^{\ast}$)\;
    $\mathcal{K}$ $\leftarrow$ update($\mathcal{K}$,~$\mathbf{\Theta}_{k}^{\ast}$)\;
    
    $\{\mathcal{O}^{\ast}_{i}\}$ $\leftarrow$ $\{\mathcal{O}^{\ast}_{i}\}$ $\cup$ $\mathcal{O}^{\ast}_{k}$\;
    }
}
$\mathcal{H}^{\ast}$ $\leftarrow$ $\mathcal{H}$\;
return $\mathcal{H}^{\ast}$, $\{\mathcal{O}_{i}^{\ast}\}$\;
\caption{Sequential grasp of multiple objects.}\label{alg:sequential_grasp}
\end{algorithm}

\subsection{One step further: greedy grasping of more objects}\label{subsec:greedy_grasping}
The proposed sequential grasping algorithm (Algorithm~\ref{alg:sequential_grasp}) enables the robotic hand to grasp multiple objects.
However, the final grasping configuration may not be ``optimal'' regarding the amount of grasped objects.
On the one hand, the solution to each grasp is not unique.
The selected mini-batch OS candidates do not guarantee optimality among all feasible OSes, and there may also exist multiple OSes that result in feasible, even optimal grasps.
On the other hand, the grasping order $\bm{\Lambda}$ affects the hand configuration for each sub-problem.
Each grasp is affected by its previous grasps and affects its subsequent grasps.
We propose the following strategies to tackle these issues in a multiple-objects grasping problem.

\subsubsection{Kinematic efficiency metric}
We measure the kinematic efficiency for each grasp configuration from three aspects:
(1) the total number of contacts, $N_c \in \mathcal{N}$,
(2) the total number of engaged finger joints, $N_q = \sum_{i=1}^{N_c}|{\bm{\theta}^{i}_{actv}}| \in \mathcal{N}$,
and (3) the capacity ratio, $\eta = {\overline{{C}}_{\mathcal{S}}}/{\|\bm{p}_{1}-\bm{p}_{2}\|} \in [1,\infty)$.
$\bm{\theta}^{i}_{actv}$ is a vector composed of all joint angles affecting the contact point on $i$th finger, and $|{\bm{\theta}^{i}_{actv}}|$ represents its cardinality.
As stated previously, we plan grasps inside opposition spaces and hence only consider grasps with two contacts.
Therefore, $N_c=2$ remains a constant value in our proposed algorithm.
$\eta$ measures the ``efficiency'' of exploiting the selected OS by taking the ratio of its maximum capacity to the actual grasp distance inside it.
A grasp that fully exploits the OS capacity has a $\eta$ value close to $1$, indicating that the size of the grasped object is slightly smaller or almost equal to the maximum capacity of the OS.
We propose the following \emph{kinematic efficiency} (KE) metric $\kappa$ by integrating the above-mentioned aspects:
\begin{equation}\label{eq:kappa}
    \kappa = e^{N_c} \cdot e^{N_q} \cdot e^{\eta}.
\end{equation}
The exponential mapping guarantees the continuity of this metric and its derivatives.
It measures the overall consumption of the kinematic redundancy in the hand model given a grasp configuration.
A more kinematically efficient grasp configuration is indicated by a smaller $\kappa$ value.
To facilitate the exploitation of kinematic redundancy, we integrate the kinematic efficiency metric into the objective function for single-object grasp (Eq.~\ref{eq:objective_function_single}) and use it as the objective function for sub-problems in multiple-objects grasp:
\begin{equation}\label{eq:objective_function_multiple}
    \mathcal{Q} = \lambda \cdot (\kappa (C_{f}+C_{t})) + (1-\lambda)\cdot C_{q},~\eta\in [0,1].
\end{equation}
This objective aims at penalizing the excessive use of opposition space capacity and improving the efficiency of utilizing kinematic redundancy for each sub-problem.
Thus, by minimizing the kinematic efficiency metric in the iterative process of grasping multiple objects, the robotic hand (1) prioritizes the OS candidate that employs fewer DoFs (e.g., the OS closer to the base of the kinematic chain), and (2) when multiple OS candidates demanding the same number of DoFs, prefers the smallest OS.
It hence preserves as much kinematic redundancy as possible for subsequent tasks during the iterative grasping process.
Notice that for the single object grasp synthesis problem, once a desired OS is given, $e^{N_c}$ and $e^{N_q}$ in $\kappa$ become constant values. The metric $\kappa$ is then only a function of $\eta$, determined by the optimization variables in contact points.
From a computational point of view, potential abrupt changes in objective function value caused by the change of employed DoFs ($N_q$) can be avoided when solving the optimization problem for each single object grasp.
We hence integrate the kinematic efficiency metric $\kappa$ in the single object grasp synthesis problem by replacing the objective function in Algorithm~\ref{alg:single_grasp} with Eq.~\ref{eq:objective_function_multiple}.

\subsubsection{Optimal greedy grasp}
The solution to a sequential grasping is deterministic, as long as the grasp sequence is defined and the size of the mini-batch OS set for each object is determined.
Therefore, to reduce the influence of grasping order on the final solution, we generate a permutation of the object's index set $\mathcal{I}$:
\begin{equation}
    \mathrm{P}_{\mathcal{I}} = \{\bm{\Lambda}_{1},\dots,\bm{\Lambda}_{N_{\Lambda}}\},~\mathcal{I}=\{1,\dots,N_{o}\},
\end{equation}
where $N_{\Lambda}=N_o!$ is the number of permutations for $N_o$ objects.
The robot then performs $N_{\Lambda}$ independent trials of sequential grasps by following each grasping order $\bm{\Lambda}$ in the permutation set $\mathrm{P}_{\bm{\Lambda}}$.
The optimal solution to the greedy grasping problem is the sequential grasp having most objects grasped; for multiple sequences result in the same amount of grasped objects, the optimal one leads to the minimum \emph{kinematic redundancy cost}:
\begin{equation}
    C_{\Lambda_{k}}^\mathrm{seq} = \sum_{i}^{N_g} {\overline{{C}}_{\mathcal{S}_{i}}},~i=1,\dots,N_{g},
\end{equation}
where $\Lambda_{k}$ is the grasping order, $\overline{{C}}_{\mathcal{S}_{i}}$ is the maximum capacity of ${\mathcal{S}_{i}}$ that is used to grasp the $i$th object.
$C_{\Lambda_{k}}^\mathrm{seq}$ describes the overall consumption of kinematic redundancy to grasp $N_g$ objects in total.
The smaller the cost, the larger the kinematic redundancy remained in the model.
We summarize our greedy grasp algorithm in Algorithm~\ref{alg:greedy_grasp}.

\begin{algorithm}[ht!]
\SetAlgoLined
\KwData{hand model $\mathcal{H}$\;
kinematic redundancy set $\mathcal{K}$\;
set of target object models $\{\mathcal{O}_{i}\}$, $i\in\mathcal{I}$\;
index set of objects $\mathcal{I}=\{1,\dots,N_{o}\}$\;}

\KwResult{hand configuration $\mathcal{H}^{\ast}$ and the set of grasped object poses $\{\mathcal{O}^{\ast}_{i}\}$, $i=1,\dots,N_g$ that associated with the optimal sequential grasp\;}

initialization\;
$\mathrm{P}_{\mathcal{I}}$ = generate\_permutation($\mathcal{I}$)\;
$p = 1$\;
\While{$p < \vert \mathrm{P}_{\mathcal{I}} \vert$}
{
    $\bm{\Lambda}_{p}$ $\leftarrow$ $\mathrm{P}_{\mathcal{I}}(p)$\;
    ($\mathcal{H}_{p}$,$\mathcal{K}$,$\{\mathcal{O}^{\ast}_{i}\}_{p}$,$C^{seq}_{\Lambda_{p}}$) $\leftarrow$ Algorithm\_2($\mathcal{H}$,$\mathcal{K}$,$\{\mathcal{O}_{i}\}$, $\bm{\Lambda}_{p}$)\;
    $p = p + 1$\;
}
$p^{\ast}$ $\leftarrow$ optimal($\arg\max_{p}(\vert\{\mathcal{O}^{\ast}_{i}\}_{p}\vert$),~$\arg\min_{p}(\{ C^{seq}_{\Lambda_{p}} \})$)\;
$\mathcal{H}^{\ast}$ $\leftarrow$ $\mathcal{H}_{p^{\ast}}$\;
$\{\mathcal{O}^{\ast}_{i}\}$ $\leftarrow$ $\{\mathcal{O}^{\ast}_{i}\}_{p^{\ast}}$\;
\caption{Greedy grasp of multiple objects.}\label{alg:greedy_grasp}
\end{algorithm}

\section{Experiments}
\begin{figure*}[htp!]
    \centering
    \includegraphics[width=\textwidth]{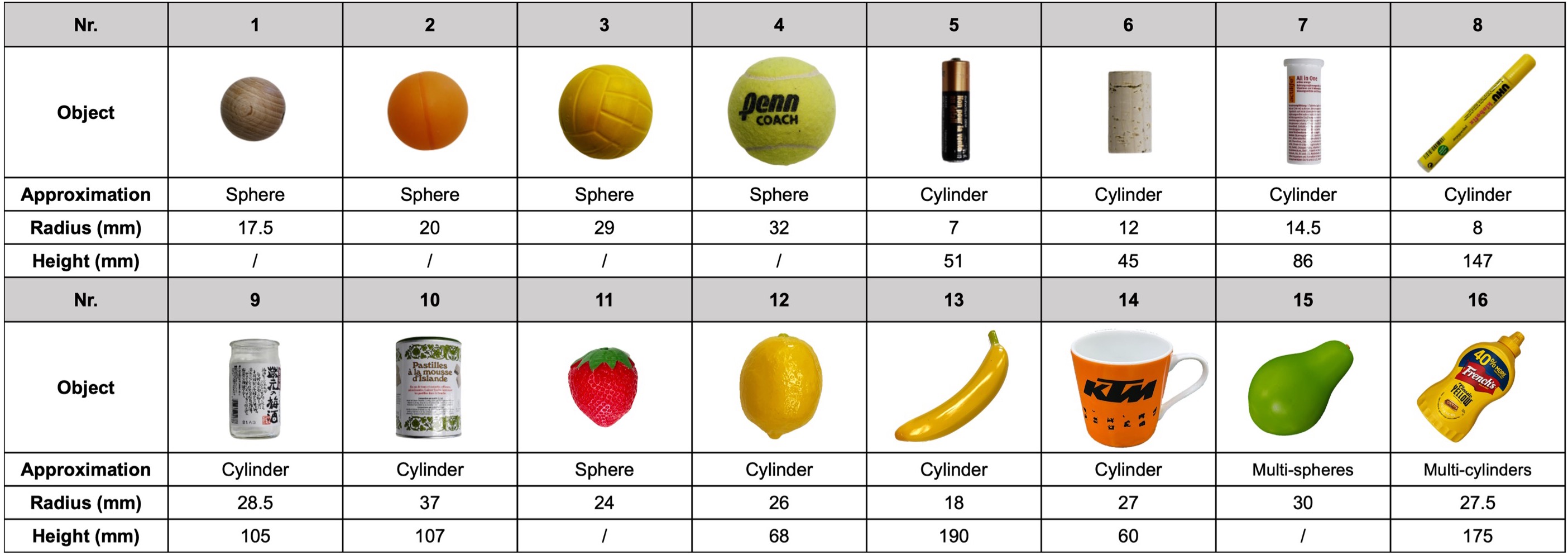}
    \caption{Experimental objects and the approximated geometric shapes.
    Objects $11$, $12$, $13$, $15$, $16$ are from the YCB-object and model set \cite{calli2015ycb}.
    The parameters of the approximated models are obtained by fitting a standard geometric shape to the object's point-cloud data. Complex object $\mathcal{O}_{15}$ is approximated using $2$ spheres (radius $17$mm and $30$mm); and $\mathcal{O}_{16}$ is approximated using $3$ parallel cylinders ($h_{1}$: $145$mm, $r_{1}$: $17.5$mm; $h_{2}$: $175$mm, $r_{2}$: $27.5$mm; $h_{3}$: $145$mm, $r_{3}$: $17.5$mm).
    For each complex object ($\mathcal{O}_{15}$ and $\mathcal{O}_{16}$), the parameters listed here are from the geometric shape that is in contact with fingers when being grasped.}
    \label{fig:objects}
\end{figure*}

We design experiments to evaluate each algorithm, respectively.
We evaluate the performance of Algorithm~\ref{alg:single_grasp} by generating grasps on different objects using both human hand and robotic hand models in various poses (Sec.~\ref{sec:single_object_grasping_experiment}), and replicate generated robot grasps on a real robotic hand.
To validate Algorithm~\ref{alg:sequential_grasp}, we generate multiple sequential grasps (Sec.~\ref{sec:sequential_grasping}).
In each trial, we randomly select three objects from the object set, generate grasp hand poses for each object in sequence, and reproduce it on a real robotic hand.
Finally, in Sec.~\ref{sec:greedy_grasping_experiments} we apply Algorithm~\ref{alg:greedy_grasp} to optimize the grasp sequence and hand configuration in a sequential grasping, enabling a robotic hand to grasp possibly many objects.

\begin{figure}[h]
    \centering
    \includegraphics[width=\columnwidth]{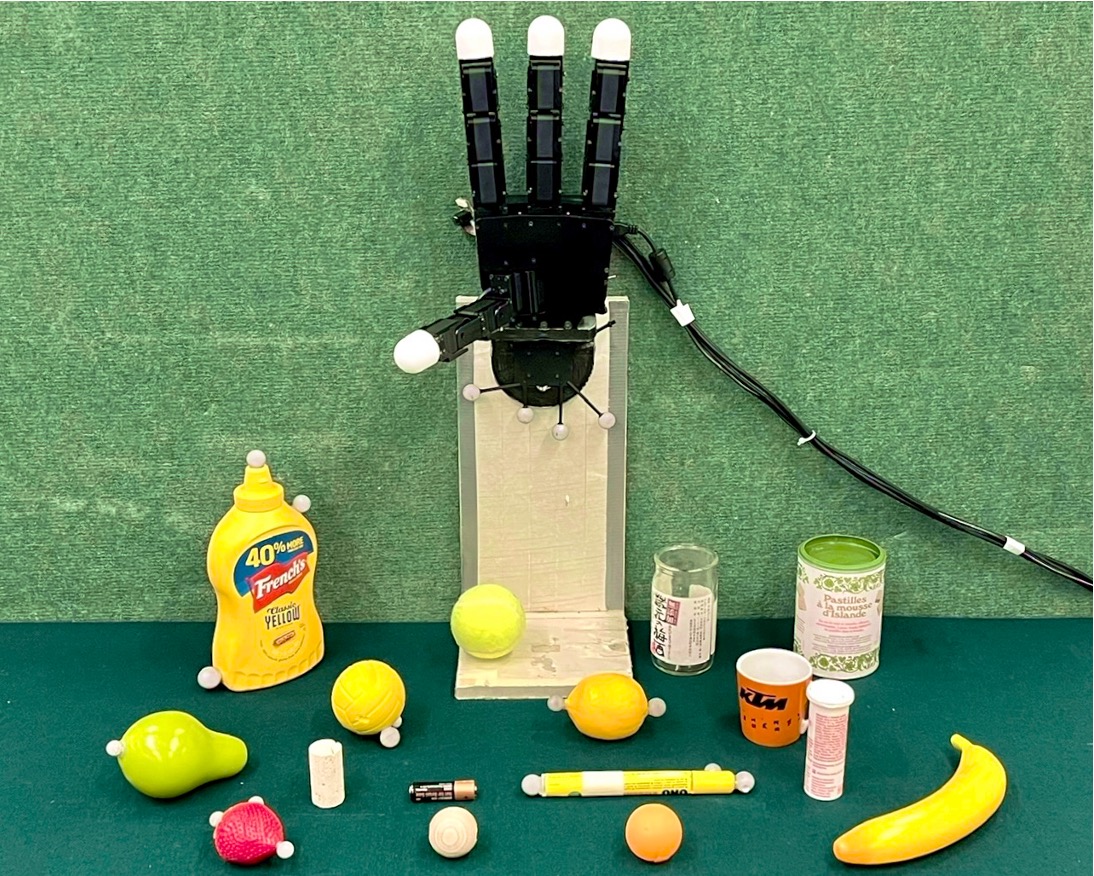}
    \caption{Experimental setup: the Allegro hand is mounted on a stand with its fingers open. Experimental objects are placed on the table.}
    \label{fig:experimental_setup}
\end{figure}

\begin{figure*}[h]
    \centering
    \includegraphics[width=\textwidth]{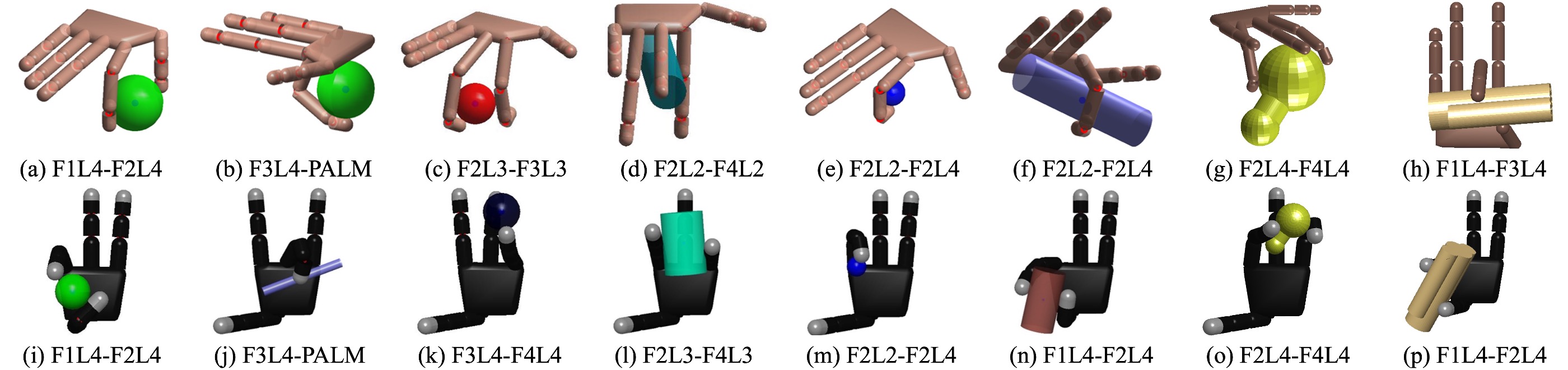}
    \caption{Examples of human-like dexterous grasping using different opposition spaces, tested on both the human hand model and the Allegro hand model.
    The title of each sub-figure indicates the OS being used to afford this grasp.}
    \label{fig:single_grasping}
\end{figure*}

\vspace{-0.1cm}
\subsection{Implementation details}
Our grasp synthesis algorithms are implemented in MATLAB (The MathWorks\textregistered, Inc.).
We use the \emph{GlobalSearch} together with the sequential quadratic programming (SQP) algorithm \cite{jorge2006numerical} in Optimization Toolbox\texttrademark, in order to find a global optimal solution\footnote{When the global optimality is not necessary, a feasible solution can be obtained faster using the \texttt{fmincon} function.}.
We formulate both the objective function and constraints as symbolic expressions using the MATLAB Symbolic Math Toolbox\texttrademark.
Prior to solving the optimization problem, we performed an offline analysis of the hand kinematic model to construct the reachability map, collision map, and opposition space set.
For simplicity, we used convex hulls to model the geometry of the reachable space for each finger phalanx.
Each friction cone is approximated by a triangular pyramid (see Appendix~\ref{app:friction_cone}).
We use a coefficient of friction $\mu = 0.5$, typical of common materials \cite{lide2004crc}.
In the objective function, we use a diagonal matrix $Q_{i}=\text{diag}[10,50,25,10]$ to prioritize the motion of each DoF (first entry corresponds to the ab-/adduction DoF) and $\lambda=0.5$ to balance costs in the grasping quality and movement economy aspects.
When formulating the constraints, each cylindrical geometry is approximated by a number of $\lceil\frac{h}{r}\rceil$ samples along the central axis, where $h$ is the height of the cylinder and $r$ is its radius.
A minimum number of $3$ samples are used in the case of $h\le r$.
The hand model moved to the fully open pose when initialized.
This configuration corresponds to the upper bound of all flexion/extension DoFs and the middle value ($0$-position) of the ab-/adduction DoFs.
Variables that parameterize object pose were randomly initialized within corresponding boundaries.

\subsection{Computational time cost}
Experiments were conducted in MATLAB R2019b on a 64-bit Ubuntu 18.04 PC with Intel(R) Core\texttrademark~i7-7700 CPU\@3.60GHz.
The average time spent on the construction and offline analyses for each hand model in simulation is summarized in Table~\ref{tab:computational_cost}.
Table~\ref{tab:file_size} reports the size of each constructed model file and associated model property files.
The time performance statistics reported here are obtained from $20$ independent experimental trials.
Modelling of the hand consists of building the kinematic chain and computing the symbolic expressions of forward kinematics of all links and joints.
The robotic hand model takes longer to construct the symbolic expressions of the kinematics, and to save the symbolic files as local MATLAB functions for future use.
This is due to the time-consuming processing of high-precision DH parameters in MATLAB symbolic expressions.
The parameters of the human hand model are rounded to integers, hence its modelling is faster despite more DoFs.
Without saving symbolic expressions, constructing the robotic model takes $2.37\pm0.04$s, and the human model takes $2.41\pm0.03$s.
The \emph{reachability map} $\{\mathcal{R}\}$ and \emph{self-collision} map $\mathcal{M}_{\mathcal{C}}$ are constructed for the entire hand model by sampling all DoFs in their motion ranges.

\begin{table}[htp]
\small\sf\centering
\caption{Time (sec) on modelling and analysis of each hand model\tablefootnote{Time consumption may vary by programming language and model representation.}.}
\label{tab:computational_cost}
\begin{tabular}{cccc}
\toprule
Model Type & Modelling & $\{\mathcal{R}\}$ & $\mathcal{M}_{\mathcal{C}}$ \\ \midrule
Human hand & $17.96\pm1.10$ & $1.27\pm0.03$ & $2.20\pm0.03$ \\
Robotic hand & $55.30\pm1.00$ & $2.67\pm0.03$ & $2.93\pm0.06$ \\ \bottomrule
\end{tabular}

\end{table}

\begin{table}[htp]
\small\sf\centering
\caption{Size (KB) of hand models and properties\tablefootnote{
Model files are saved in \texttt{mat} format.}.}
\label{tab:file_size}
\begin{tabular}{cccc}
\toprule
Model Type & Model file & $\{\mathcal{R}\}$ file & $\mathcal{M}_{\mathcal{C}}$ file \\ \midrule
Human hand & $635.7$ & $306.3$ & $0.486$ \\
Robotic hand & $547.3$ & $627.3$ & $0.456$ \\ \bottomrule
\end{tabular}
\end{table}

Formulating a single object grasp synthesis problem takes less than a minute, resulting in $20-30$ optimization variables, $100-200$ nonlinear inequality constraints, and around $20$ nonlinear equality constraints.
Such an optimization problem can be solved usually within seconds (see Fig.~\ref{fig:SG_time_analysis}(a), first column, for a detailed comparison of \emph{analysis time} and \emph{solver time} in a single grasp).

\subsection{Robotic experimental setup}
We reproduced our synthesized grasps on a 16-DoF real Allegro robotic left hand (Wonik Robotics Co., Ltd.).
The robotic hand was mounted on a metal adapter on a table surface in a standing posture (see Fig.~\ref{fig:experimental_setup}).
The hand was connected to an Ubuntu 18.04 desktop PC through a CAN bus.
It was controlled by a joint position PID controller in ROS Melodic at $10$ Hz.
We selected $16$ everyday objects (see Fig.~\ref{fig:objects}) for robotic experiments, and the geometry of each object has been approximated using one or multiple sphere(s) or cylinder(s) of various sizes.
Objects $11$, $12$, $13$, $15$, $16$ in the list are from the YCB object set \cite{calli2015ycb}.
OptiTrack\texttrademark motion capture system was used to localize the robotic hand and each target object in the experiment.
Reflexive markers were affixed to the support base of the robotic hand and the surface of each object, to track their positions in space.

\subsection{Human-like dexterous grasping of single object using arbitrary surface regions}\label{sec:single_object_grasping_experiment}
We validated our proposed human-like dexterous grasping Algorithm~\ref{alg:single_grasp} by generating grasps of various objects using different opposition spaces in the hand model.

\subsubsection{Simulation results}
We first generated grasps using spherical and cylindrical objects of different sizes on both the human right-hand model and the Allegro left-hand model in simulation (Fig.~\ref{fig:single_grasping}).
For each hand model, we validated our algorithm by generating grasps using different pairs of opposing regions.
Such grasp poses include (1) the \emph{pinch grasp}, using either the commonly seen thumb-index (or middle) coupling ((a), (h), (i), (n), and (p)) or the combination of finger and palm ((b), (f), and (j)); (2) the \emph{wrap grasp} that uses phalanges from one single finger ((e) and (m)); (3) the \emph{addiction grasp} by the lateral faces of finger phalanges, regardless of whether fingers are adjacent ((c) and (k)) or not ((d), (g), (l), and (o)).
Except pinch grasps that employ the thumb-tip and the index / middle fingertips, most of these grasps are atypical for robotic hands but are common for humans, and have been documented in human grasp taxonomy \cite{feix2015grasp}.
Moreover, it is worth noting that in the pinch grasp formed by the thumb and the index finger ((i) and (n)), the thumb distal phalanx established contact with the object surface on its ``back'' surface, seems unnatural compared to a human grasp.
This is mainly because the distal phalanges have been modeled as cylinders without discriminating the frontal and back faces.
It is, however, possible to force the contacts to appear at a desired region of a finger phalanx by modifying the boundary constraints of contact coordinates (see Appendix~\ref{app:contact_model}).

\subsubsection{Evaluation on robotic hand}
\label{sec:evaluation_on_robotic_hand}
\begin{figure*}[htp!]
\centering
    \includegraphics[width=\textwidth]{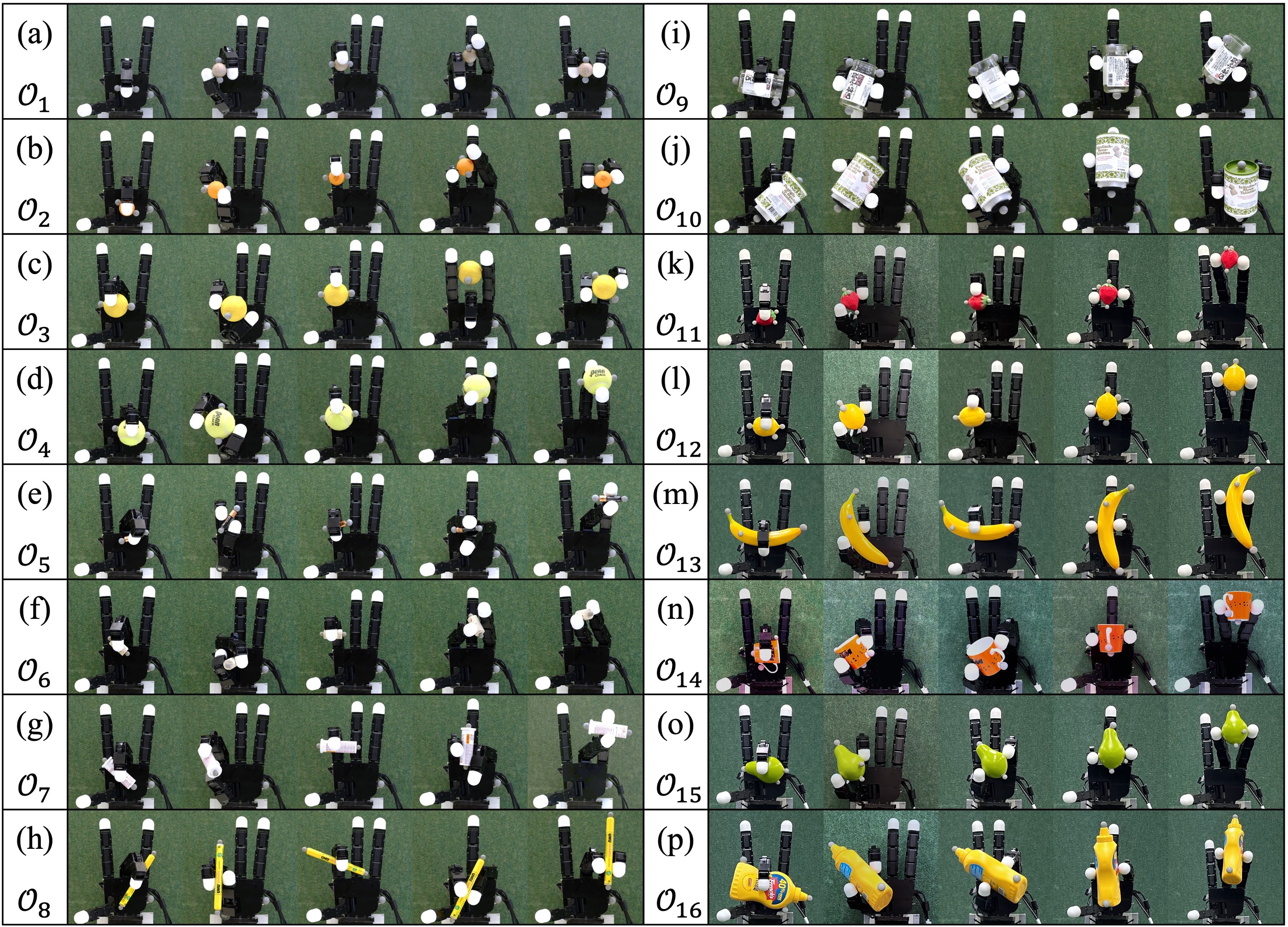}
    \caption{The robotic hand replicates dexterous grasps of each experimental object using various opposition spaces spanned by arbitrary opposing surface regions.
    Each sub-figure from $(a)$ to $(p)$ corresponds to the objects from number $1$ to $16$ in Fig.~\ref{fig:objects}.}
    \label{fig:SG}
\end{figure*}

Each grasp was first generated in simulation by applying Algorithm~\ref{alg:single_grasp}.
Once a successful grasp has been obtained, the human experimenter held the target object to approach its desired spatial pose ($\mathcal{O}^{\ast}$ in Algorithm~\ref{alg:single_grasp}).
The robotic hand is controlled by joint position command $q_{cmd}$ (sent at $10$Hz in experiments):
\begin{equation}
    q_{cmd} = q_{des} - K_{e}~(q_{des}-q_{0}),
\end{equation}
where $q_{0}$ is the initial joint position, $q_{des}$ is the joint position in the desired grasp pose, and $K_{e}$ is the error feedback gain:
\begin{equation}
    K_{e} = k \cdot \lVert \mathbf{o}-\mathbf{o}_{des} \rVert + (1-k)\cdot(1-\langle \bm{q}^{\mathcal{O}},~\bm{q}^{\mathcal{O}}
    _{des}\rangle^{2})
\end{equation}
with $\lVert \mathbf{o}-\mathbf{o}_{des} \rVert$ being the Euclidean distance between the current object center position $\mathbf{o}$ and desired position $\mathbf{o}_{des}$, and $\langle \bm{q}^{\mathcal{O}},~\bm{q}^{\mathcal{O}}_{des} \rangle$ the quaternion distance between object's current orientation $\bm{q}^{\mathcal{O}}$ and desired orientation $\bm{q}^{\mathcal{O}}_{des}$; $k$ balances the error in object position and orientation ($k=0.5$ in our experiments).
After the robotic hand had grasped the object, the human experimenter released the object and checked if the grasp was stable.
If the object cannot be grasped by the desired links, or if the object has slipped out of hand, the grasp is considered unstable.
In this case, the robotic hand returned to its initial configuration, increased (or decreased) the corresponding joints in $q_{des}$ by $\epsilon$ (we used $\epsilon = 5\%$ in our experiments), such that the opposing hand links in the grasp moved closer against each other to increase the contact force.
In a stable grasp, the robotic joint angles and object spatial positions have been registered as the initial configuration for planning the subsequent grasp.
If the object cannot be stably grasped after adjusting finger joints, the experimental trial is considered a failure.
We generated grasps of each selected object using five distinct grasp poses: (1) pinch grasp by a finger and the palm, tested on F3L4-PALM, (2) pinch grasp using the thumb and the index finger, tested on F1L4-F2L4, (3) wrap grasp, tested on F2L2-F2L4, (4) adduction grasp using non-adjacent fingers, tested on F2L3-F4L3 or F2L4-F4L4, and (5) adduction grip using adjacent fingers, tested on F3L4-F4L4.
Fig.~\ref{fig:SG} lists all grasps performed by the real robotic hand.
Notice that for $\mathcal{O}_{9}$ (Fig.~\ref{fig:SG}(i)), $\mathcal{O}_{10}$ (Fig.~\ref{fig:SG}(j)), $\mathcal{O}_{14}$ (Fig.~\ref{fig:SG}(n)), $\mathcal{O}_{15}$ (Fig.~\ref{fig:SG}(o)), and $\mathcal{O}_{16}$ (Fig.~\ref{fig:SG}(p)), the wrap grasp cannot be achieved, because the OS F2L2-F2L4 is not geometrically permissive for such large objects.
Then F1L4-F4L4 was employed instead, resulting in pinch grasp poses.
Moreover, the plastic banana $\mathcal{O}_{13}$ (Fig.~\ref{fig:SG}(m)) could hardly be stably grasped using F3L4-PALM, because the object surface is smooth and angular, making it difficult to maintain a stable contact.
Instead, a wrap grasp by the middle finger was used.
Most objects can be successfully grasped at the calculated grasp configuration.
A few grasps were unstable and slight adjustment has been made in re-grasp.
To evaluate the deviation between simulation results and grasps by a real robotic hand, we quantitatively analyzed registered robotic joint angles and object positions.

\begin{figure*}[ht]
    \centering
    \includegraphics[width=\textwidth]{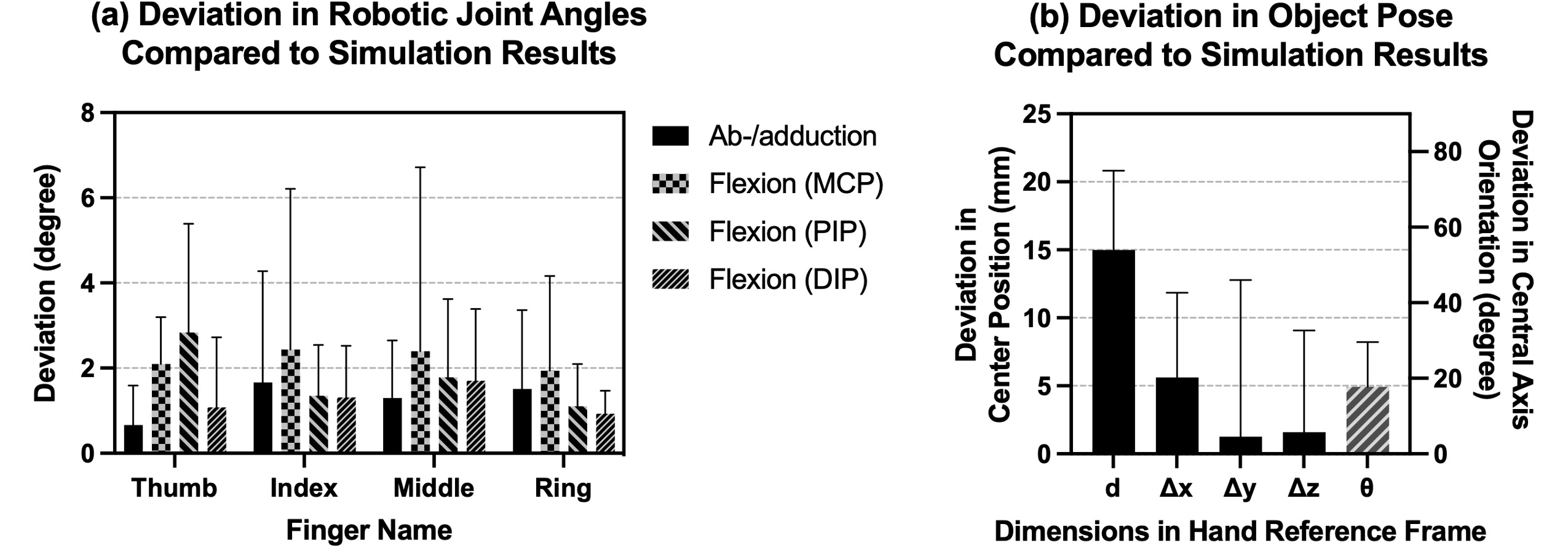}
    \caption{Deviation in parameters between real robotic grasps and simulated results.
    (a) Absolute values of deviation in robot joint angles (in degree).
    (b) Absolute values of deviation in grasp object position (mm, left y-axis) and orientation (deg, right y-axis). $d$ is the spatial distance between grasped real object center and simulated center, $\Delta x$, $\Delta y$, $\Delta z$ denotes deviation in each dimension, respectively. Deviation in orientation is calculated as the included angle $\theta$ between object quaternion $\bm{q}^{\mathcal{O}}$ and calculated quaternion $\bm{q}^{\mathcal{O}}_{des}$ as $\theta=\arccos(2\langle\bm{q}^{\mathcal{O}},~\bm{q}^{\mathcal{O}}_{des} \rangle^{2}-1)$.}
    \label{fig:experiments_simulation_deviation}
\end{figure*}

We first calculated the deviation in joint angles across all single object grasp trials, as the absolute difference between simulated joint angle values and the recorded robot joint angles while replicating the grasp.
For most joints (Fig.~\ref{fig:experiments_simulation_deviation}(a)), the deviation is smaller than $5$ degrees. The largest deviation of $24.56$ degree was observed in the second joint (the MCP flexion/extension DoF) of the index finger when the robotic hand grasped $\mathcal{O}_{4}$ by wrapping.
We also compared the deviation in object positions between simulation results and robotic experiments.
Fig.~\ref{fig:experiments_simulation_deviation}(b) illustrated the absolute deviation in spatial distance and orientation.
Across all experimental trials, objects in the robotic experiments differed from the simulated position by $d = 14.98\pm 5.84$mm (mean $\pm$ std) and deviated in orientation by $\theta = 17.72\pm 11.85$deg.

Our robotic grasping experimental results demonstrate that the robotic hand could grasp the target object using configurations generated by applying Algorithm~\ref{alg:single_grasp}.
The deviations in joint angles and object positions are mainly due to the noise and also the inaccuracy in the modelling of the robotic hand.
For example, the phalanges of the robotic hand have been approximated by cylinders for simplification, but the real Allegro hand's phalanges are in fact cuboid.

\subsection{Sequential grasping of multiple objects}\label{sec:sequential_grasping}

\begin{figure*}[htp!]
    \centering
    \includegraphics[width=\textwidth]{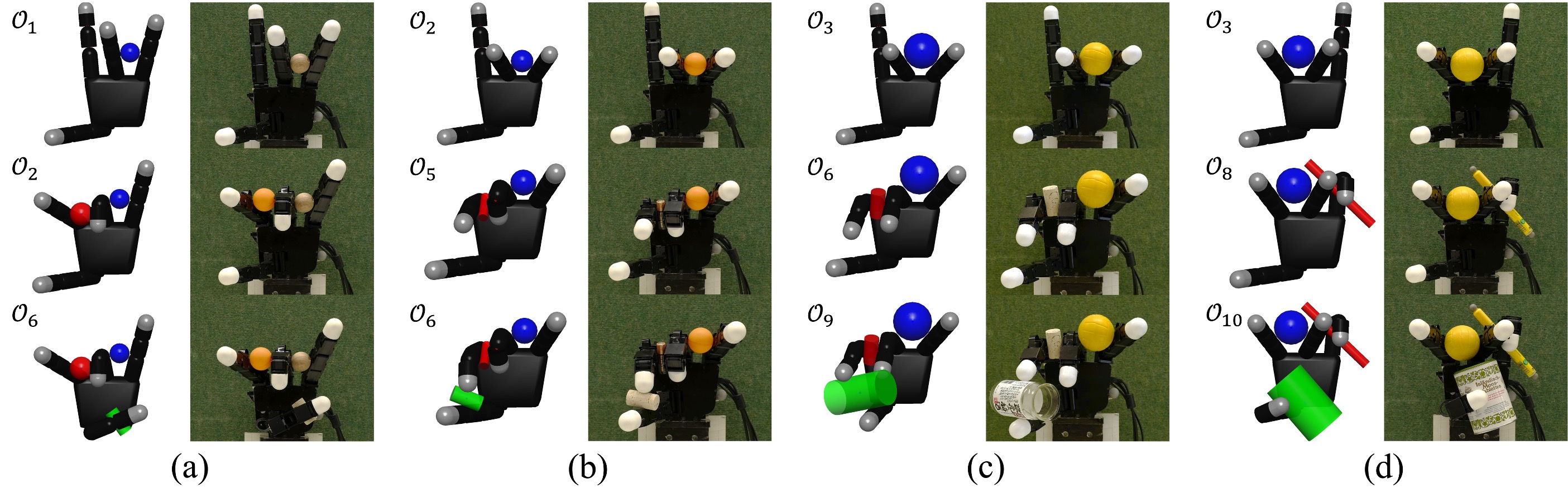}
    \caption{Examples of sequential grasping of three randomly selected objects.
    Grasp configuration at each step was generated by following Algorithm~\ref{alg:sequential_grasp}.
    Each sub-figure demonstrates one independent trial of sequential grasp from top to bottom: the left column shows the simulated grasp configuration, and the right column shows the real grasp performed by the robotic hand.}
    \label{fig:MG}
\end{figure*}

\begin{figure*}[ht!]
    \centering
    \includegraphics[width=\textwidth]{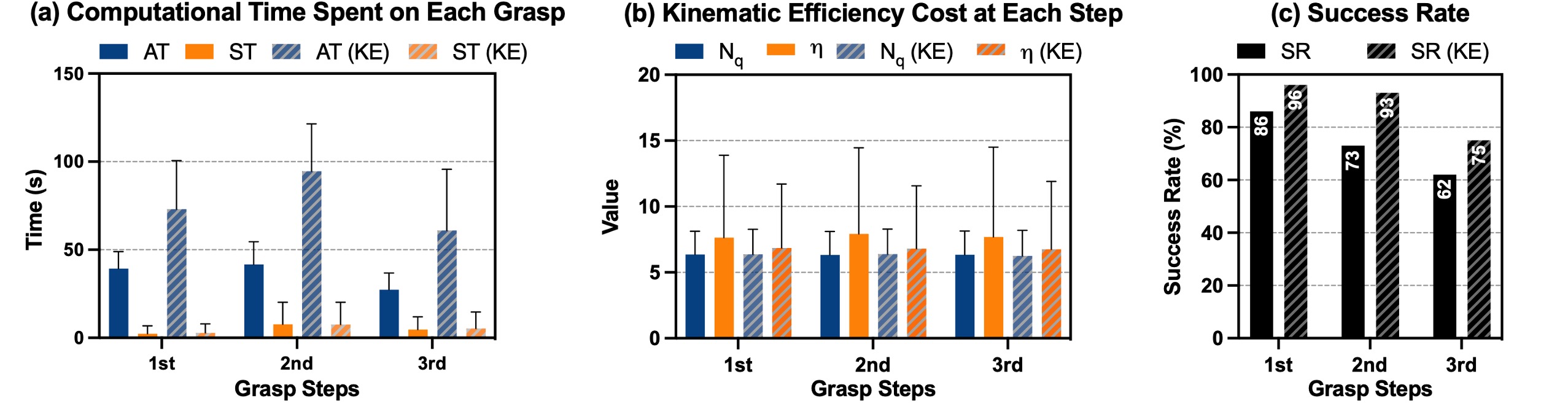}
    \caption{Sequential grasping experimental performance by applying Algorithm~\ref{alg:sequential_grasp}.
    One object is grasped at each grasp step.
    ``KE'' indicates using objective function Eq.~\ref{eq:objective_function_multiple} instead of Eq.~\ref{eq:objective_function_single} to optimize the kinematic efficiency metric at each grasp step.
    (a) The total computational time spent on generating the final grasp of each object (including the total time spent on evaluating all OS candidates in the case of ``KE'').
    \emph{AT} indicates analysis time, including the time spent on modelling, constructing/updating $\{\mathcal{R}\}$ and $\mathcal{M}_{\mathcal{C}}$.
    \emph{ST} stands for solver time, i.e., the time taken to solve the optimization using the SQP algorithm in the \texttt{fmincon} function in MATLAB (Line 6 in Algorithm~2, and Line 11 in Algorithm~1, correspondingly).
    (b) Comparison of kinematic efficiency cost at each step. $N_{q}$ is the number of engaged hand DoFs, $\eta$ is the capacity ratio of the used OS in this grasp.
    (c) The success rate (SR) of obtaining a successful grasp at each step. The number on each bar is the success rate in percentage.}
    \label{fig:SG_time_analysis}
\end{figure*}

We validated our sequential grasping algorithm in four trials.
In each trial, three objects have been randomly selected from our object list (Fig.~\ref{fig:objects}).
For each object, three OS candidates have been tested, and the one that optimizes the objective function (Eq.~\ref{eq:objective_function_multiple}) has been assigned the final grasp configuration.
Following our proposed Algorithm~\ref{alg:sequential_grasp}, we first generated the optimal sequential grasp configurations for target objects in ascending order of the object ordinal number.
The optimal solution at each step minimizes the use of the kinematic capacity of hand among all selected OS candidates for the object.
At each step, once an object has been successfully grasped, both the joint configuration and object position have been recorded for reproducing grasps on a real robotic hand afterward.
We see in Fig.~\ref{fig:MG} (left panel of all sub-figures) that successful grasping sequences can be generated by solving the problem in all trials.
Then, we reproduced the simulated sequential grasps on the real robotic hand, following the same procedure in Sec.~\ref{sec:evaluation_on_robotic_hand}.
Experimental results (see Fig.~\ref{fig:MG}, right panel of each sub-figure) demonstrate that the robotic hand was able to perform successful sequential grasping of objects in all trials.

To evaluate the overall performance of our sequential grasp algorithm (Algorithm~\ref{alg:sequential_grasp}), and also to validate the use of our kinematic efficiency metric in the objective function, we simulated 100 trials of sequential grasp in each of the two experimental conditions: (1) \emph{regular sequential grasp} (use Eq.~\ref{eq:objective_function_single}), and (2) \emph{KE sequential grasp} with optimizing kinematic efficiency (use Eq.~\ref{eq:objective_function_multiple}).
Results are summarized in Fig.~\ref{fig:SG_time_analysis}. Performance metrics with ``KE'' indicate the integration of optimizing kinematic efficiency.
In each condition, 3 objects were randomly selected from the object list (see Fig.~\ref{fig:objects}) at the beginning of each trial.

In the \emph{regular sequential grasp} condition, an OS was \emph{randomly} selected from the OS set of the object, and then used to plan for the grasp at each step.
The randomly selected OSes were geometrically permissive for the 1st and 2nd objects in all $100$ trials; however, in $5$ trials, no more feasible OS can be found for the 3rd object.
Grasp 1 and 2 took approximately the same total time to analyze and solve the problem, and the 3rd grasp was slightly faster in comparison (see Fig.~\ref{fig:SG_time_analysis}(a)).
Kinematic efficiency cost did not vary much across the grasp steps (Fig.~\ref{fig:SG_time_analysis}(b)). In the first grasp, the average involved finger DoFs $N_{q}$ is (mean$\pm$std) $6.36\pm1.76$; and $N_{q}$ is $6.32\pm1.79$ for the 2nd and $6.34\pm1.80$ for the 3rd grasp, respectively. The capacity ratio $\eta$ also remains similar for each grasp: $\eta_{1}:~7.64\pm6.26$, $\eta_{2}:~7.93\pm6.52$, and $\eta_{3}:~7.69\pm6.82$.
The overall success rate is $86.0\%$ for the 1st grasp, $73.0\%$ for the 2nd grasp, and $62\%$ for the 3rd grasp (Fig.~\ref{fig:SG_time_analysis}(c)).
This leads to $74$ successful grasp sequences of at least two objects, and $43$ sequences of successfully grasping all three objects out of the $100$ experimental trials.
Moreover, among the $295$ objects grasped in the $100$ trials, $116$ were spheres (incl. $\mathcal{O}_{15}$), and $179$ were cylinders (incl. $\mathcal{O}_{16}$).
Spheres took shorter solver time ($1.26\pm 1.10$s) with a higher success rate of $82.76\%$, compared to cylinders ($8.93\pm 15.57$s) at a lower success rate of $68.16\%$.

In the \emph{KE sequential grasp} case, at each step, three OSes were randomly selected from the geometrically permissive OS set and used to plan for grasps. The OS that resulted in the most kinematic-efficient grasp has been selected as the final grasp at each step.
Geometrically permissive OSes were successfully selected for the first two grasps in all trials.
For the 3rd object, no feasible OSes could be found in $8$ trials in total.
With such an optimization strategy, KE-sequential grasp generated less kinematic efficiency cost (Fig.~\ref{fig:SG_time_analysis}(b)), particularly, smaller $\eta$ values at every grasp step ($\eta_{1}:~6.85\pm4.86$, $\eta_{2}:~6.80\pm4.78$, and $\eta_{3}:~6.75\pm5.16$).
This contributed to the improvements in success rate for all steps (Fig.~\ref{fig:SG_time_analysis}(c)).
Out of the $100$ trials, at least two objects were successfully grasped in a total of $89$ trials, and all three objects were grasped in $61$ trials.
This also improved the success rate of grasping both spherical objects ($115$ in total, success rate $96.52\%$, time $1.32\pm0.97$s) and cylindrical objects ($177$ in total, success rate $83.05\%$, time $7.86\pm11.49$s).
The price paid was the longer AT due to the need to analyze more OSes (Fig.~\ref{fig:SG_time_analysis}(a)), but the ST did not differ much from the other experimental case.

\subsection{Greedy grasping of multiple objects}\label{sec:greedy_grasping_experiments}
\begin{figure*}[htp!]
    \centering
    \includegraphics[width=\textwidth]{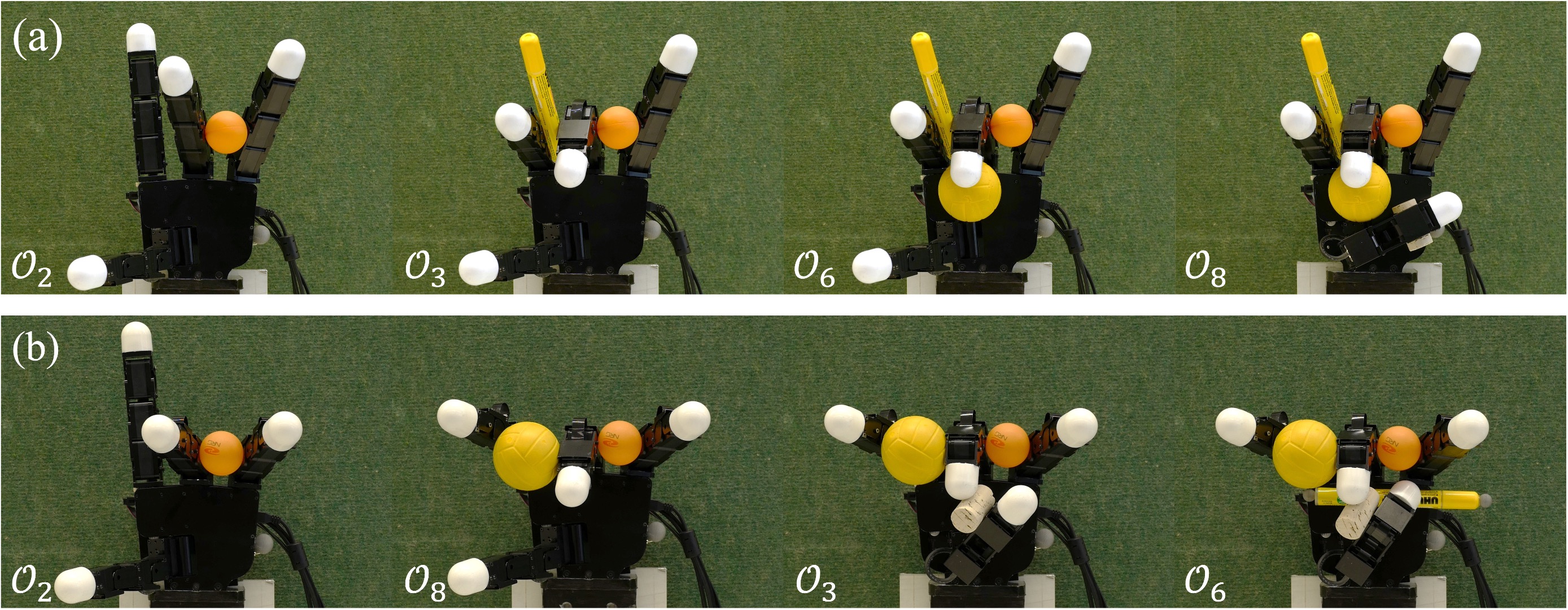}
    \caption{Greedy grasp of multiple objects using robotic hand following the proposed Algorithm~\ref{alg:greedy_grasp}. Grasps (a) $\mathcal{O}_2\rightarrow\mathcal{O}_3\rightarrow\mathcal{O}_6\rightarrow\mathcal{O}_8$ and (b) $\mathcal{O}_2\rightarrow\mathcal{O}_8\rightarrow\mathcal{O}_3\rightarrow\mathcal{O}_6$ followed different grasping orders, both led to successful solutions.}
    \label{fig:GG}
\end{figure*}

We demonstrate in Fig.~\ref{fig:MG} that the robotic hand was able to grasp three objects, and demanded three opposition spaces spanned by $6$ virtual fingers in total.
In this scenario, we evaluate our proposed greedy grasp method (Algorithm~\ref{alg:greedy_grasp}) by further exploiting the kinematic redundancy of the hand and trying to grasp more objects.
For a four-finger robotic hand used in this study, fingers must be shared in different grasps.
It hence demands proper allocation of every single object grasped within the kinematic structure of the hand in the iterative process.
We selected four objects out of the list for evaluation: $\mathcal{O}_{2}$, $\mathcal{O}_{3}$, $\mathcal{O}_{6}$, and $\mathcal{O}_{8}$.
The selection of objects took the geometric size of objects into consideration.
For example, large objects ($\mathcal{O}_{9}$ and $\mathcal{O}_{10}$) may occupy too much space inside the hand kinematic structure and thus lead to a lesser chance of a successful result.
The permutation of selected objects' index set $\mathcal{I} = [2,3,6,8]$ results in $24$ grasping orders.
For each object, three OS candidates were provided for testing.
Following our proposed algorithm, the robotic hand successfully grasped all four objects in two different poses.
Fig.~\ref{fig:GG} shows the generated grasping sequences performed by the robotic hand.
In Fig.~\ref{fig:GG}(a), the robotic hand followed the same object order as in Fig.~\ref{fig:MG}(a) for the first three objects.
When grasping the third object $\mathcal{O}_{6}$, the robotic hand employed the opposition space F1L3-F3L4, instead of F1L4-PALM that has been used in Fig.~\ref{fig:MG}(a).
This choice spared the space of the palm, thus enabling one extra grasp of $\mathcal{O}_{8}$, using the space of F1L4-PALM.
Grasp in Fig.~\ref{fig:GG}(b) followed a different order, but also resulted in a successful grasping sequence.
The most commonly violated constraints observed in failure trials are collision-free constraints and stability constraints (force-closure properties).
A total number of $8$ contact points are required to span four opposition spaces, hence reusing finger is inevitable.
In both successful trials, multiple links (including the palm) have been involved in more than one grasp.
In particular, the middle finger has been shared in three grasps in both cases.
The thumb has been reused to grasp $\mathcal{O}_{6}$ and $\mathcal{O}_{8}$ in Fig.~\ref{fig:GG}(a), while the palm surface has been shared in the grasp of $\mathcal{O}_{3}$ and $\mathcal{O}_{6}$ in Fig.~\ref{fig:GG}(b).
Experimental results demonstrate that our proposed Algorithm~\ref{alg:greedy_grasp} enables the hand to grasp more objects than in Sec.~\ref{sec:sequential_grasping} by further exploiting its kinematic redundancy.

\section{Discussion}
We have demonstrated that by exploiting the kinematic redundancy, a robotic hand with sufficient DoFs can dexterously grasp multiple objects in sequence and hold multiple grasped objects simultaneously.
Such grasp types are no longer limited to the common use of only the fingertips and the inner surface of the hand.
Instead, our human-like dexterous grasp synthesis algorithm enables the employment of arbitrary opposing surfaces of hand kinematic structure to achieve atypical grasp types, which are common for humans but rarely seen in robotic tasks.
Our proposed approaches largely increase the dexterity of the robotic hand and hence empower the hand to grasp multiple objects.
We validated our proposed approaches on both a 20-DoF human hand model and a 16-DoF robotic hand model, and also on a real robotic hand.
Experimental results verified the efficiency of our proposed algorithms in both single-object grasping and multi-object grasping experiments.
Our grasp synthesis algorithm has a hierarchical formulation that combines \emph{high-level task-space planning}, i.e., the analysis of kinematic capacity in Cartesian space and the selection of opposition space, and \emph{low-level joint-space planning}, i.e., the constrained optimization of the search for a feasible solution.
This hierarchical formulation has multiple advantages in comparison to state-of-the-art approaches.
First, it alleviates the computational demands for planning in task space.
Constructing the graspability map \cite{roa2011graspability} for a given object demands sampling hand spatial locations and orientations around the target object with desired properties such as collision-free and stability being validated for every sample.
Sampling such numerous parameters inevitably entails a huge computational cost, and the quality of results largely relies on the number of samples.
In contrast, our proposed approach divides the above-mentioned procedure into task-space planning and joint-space planning.
We formulate the problem in a hand-centered instead of an object-centered point of view, and only use the task-space information (the reachability map) to roughly restrict the relative position and orientation of the target object with respect to the robotic hand.
The optimization in joint space determines the specific grasp configuration and guarantees the constraints are satisfied.
Likewise, our approach can be applied to plan the grasps with particular configurations, such as particular hand poses, object poses, or contact regions. This can be achieved by introducing constraints on the parameters or expressions describing the hand pose, the object pose, or the contact.
Our proposed approach does not require voxelization of the spatial volume, and a precise solution in continuous space can be obtained.
Moreover, the properties of the hand kinematic model obtained in task-space analysis can be reused in novel tasks.
Second, the commonly used human-like reach-and-grasp motion by simply closing robotic fingers to grasp an object is difficult to ensure grasp quality for finger wrap grasp or adduction grasp. Generating these atypical grasp types are demanded in specific tasks.
For example, \cite{el2013generation} proposed a solution to tackle this problem by generating a library of potential grasps based on sampling.
This is, however, inefficient for satisfying a particular task demand, and also requires large computational costs.
We provide an alternative solution by exploiting the existing opposing spaces in the kinematic structure.
The task-space analysis in our approach efficiently determines the pair of links that can be employed to generate a task-demanded grasp type.
For example, selecting a pair of links from adjacent fingers will naturally result in an adduction grasp; and generating a finger wrap grasp only needs to select an opposition space constructed by links from the same finger.
Our human-like dexterous grasp synthesis algorithm enables the generation of all opposition grasps that have been discovered in human grasping hand poses \cite{iberall1986opposition}, and also avoids the computational costs caused by sampling.
Third, by combining the analysis in both task space and joint space, our proposed approach enables the reuse of analysis results obtained in previous tasks, hence reducing the computation demands.
In particular, collisions among hand links and objects must be avoided to achieve a successful grasp.
Such constraints are commonly handled in the literature by either checking for all feasible poses of the hand in task space or searching in entire joint spaces for collision-free configurations.
We formulate collision constraints by analyzing the hand's reachability map.
Our constructed self-collision map reveals that potential collisions exist only among specific pairs of links.
Therefore, hand links that do not collide in Cartesian space are excluded from grasp synthesis.
Furthermore, we also reduce the number of involved DoFs by selecting the desired OS candidate prior to joint-space optimization.
OSes that are not geometrically permissive for the target object along with their associated finger DoFs are removed.
Moreover, it is worth noting that our approach provides the potential to independently manipulate each grasped object.
As each object is grasped by at least two contacts, when using different groups of fingers, the multiple grasped objects are completely independent of each other (e.g., grasp one object with the thumb and index finger, and another with the middle finger and ring finger).
In this case, it is easy to perform independent manipulation of each object.

It is worth noting that the proposed geometrically permissive condition is sufficient but not necessary for an OS to result in a feasible grasp.
This is because it does not guarantee that the multiple constraints of the problem can be satisfied, especially when the constraints are also related to the specific parameters of the problem, for example, the coefficient of friction.
Nevertheless, performing such geometric analysis in task space helps us to eliminate irrelevant variables and only keep those that are necessary to the problem, thus reducing the complexity of the low-level joint space optimization.
Moreover, global optimality is hard to obtain in the planning of multiple objects grasping, mainly due to the fact that multiple geometrically permissive OSes may exist in high-level planning.
The selection of OS affects not only the current grasp, but also subsequent grasps.
The only way to guarantee global optimality is to iterate through all feasible OSes for each target object; however, this will result in a significant increase in the dimensionality of the problem, as the number of sub-problems is the cumulative multiplication of feasible OSes for each object.
Therefore, we balance the optimality and problem complexity by prioritizing the most kinematically efficient OSes on the one hand, and testing a mini-batch of OS candidates on the other hand.
We have validated our proposed approaches by considering spherical and cylindrical objects.
Nevertheless, our algorithms are applicable in general cases and the scope of application should not be limited by the models of robots or objects used.
Our approach can be applied, as long as kinematic redundancy exists in the robot model.
The geometric shape of an irregular-shaped object can be simply approximated by a composition of elementary geometries, such as multiple spheres \cite{hubbard1996approximating}.
Modelling inaccuracy and uncertainty can be compensated by the robotic controller, for example, impedance control, or force regulation based on tactile feedback.
It is also feasible to construct an implicit model of object geometry by sampling data on the object's surface offline \cite{el2013generation}.
Implicit geometry models can be represented using learning approaches such as the support vector regression model or the Gaussian process regression model.
This shall be considered in future work as an extension of our current approaches.
Furthermore, we implemented our algorithms in MATLAB, as we focused more on the fast development and analysis of algorithms, rather than on improving the computational performance. In the future, our algorithm can be potentially integrated into grasping simulation platforms, such as \emph{GraspIt!} \cite{miller2004graspit}. Moreover, the performance of our algorithms can also be substantially improved with more efficient programming languages and using more efficient optimization solvers.

\section{Conclusion}
This study aimed at improving the dexterity of robotic hands in grasping tasks by using all the degrees of freedom of the hand.
To this end, we developed a framework that uses all surface regions on the fingers and the palm as support to simultaneously maintain a grasp of multiple objects.
Our approaches extend state-of-the-art research in robotic grasping in three aspects.
First, to our knowledge, this is the first example of an approach to enable sequential grasping of multiple objects by a single robotic hand, while holding already grasped objects.
Second, the grasp is no longer restricted to using fingertips and the inner surfaces of the hand.
As long as a pair of opposing regions exists in the hand model and spans an opposition space, it can be employed to synthesize a grasp. While we have not considered all surfaces, the approach is not limited to the surface considered here and could be extended to all hand surface regions.
Third, we propose a kinematic efficiency metric as a quantitative measure of the kinematic redundancy in the robot model. On this basis, we proposed an associated strategy that facilitates the exploitation of kinematic redundancy in the planning of sequential grasping tasks.
Experimental results demonstrated that the robotic hand was able to successfully perform dexterous grasps of a variety of everyday objects, and also to grasp and hold multiple objects, with only slight deviations from the simulated grasps.
The algorithms proposed in this study are not restricted to the hand models or object models used and can be easily applied to any robotic hands or manipulators having multiple DoFs.
Our approach of exploiting kinematic redundancy has largely improved the dexterity of multi-DoF robotic hands in grasping tasks.
It offers the prospect of designing robot dexterous manipulation algorithms, and may also inspire the design of novel robotic hands and manipulators.

\begin{appendices}
\section{Model of contact on cylindrical geometry}\label{app:contact_model}
\begin{figure}[!ht]
\centering
    \includegraphics[width=\columnwidth]{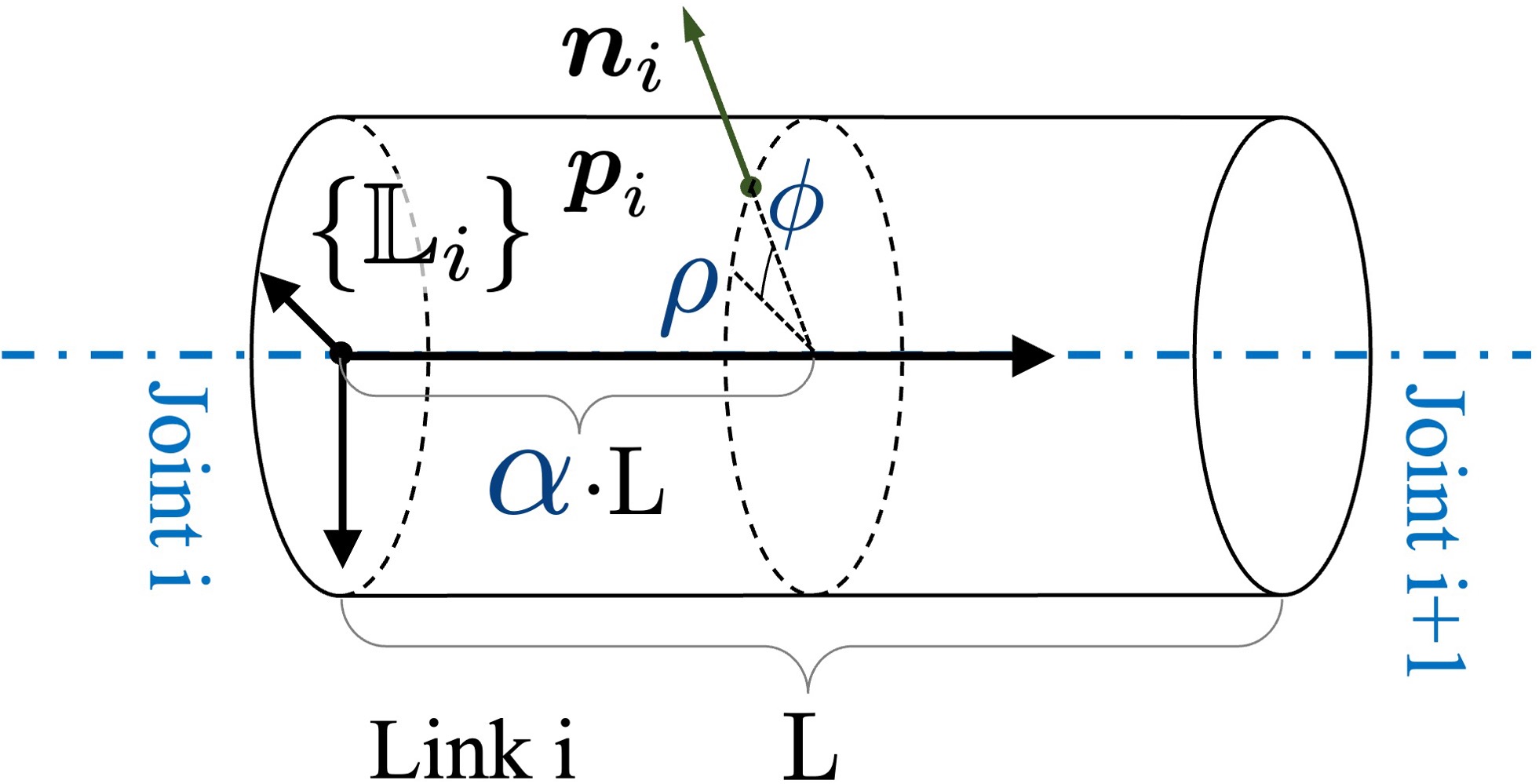}
    \caption{Parameterization of a contact point $\bm{p}_{i}$ on a finger link $\mathcal{L}_{i}$.}
    \label{fig:cylindrical_coordinates}
\end{figure}
A contact point $\bm{p}_{i}$ on an arbitrary finger phalanx $\mathcal{L}_{i}$ is parameterized by its cylindrical coordinates (Fig.~\ref{fig:cylindrical_coordinates}): (1) $\alpha$, the height ratio from the link base to the contact in the local link reference frame, (2) $\phi$, the angular coordinate, and (3) $\rho$, the radical distance of the contact point to the central axis of the link.
We represent the height of $\bm{p}$ as a ratio $\alpha\in[0,1]$ of the total link length ${L}$, the height is thus calculated as $\alpha {L}$.
We enable the contact to happen at any point on the link surface, thus $\phi\in(-\pi,\pi]$; $\rho$ equals the radius of the link.
Contact on the hand palm can only happen on the inner surface of the palm, which is modeled as a rectangular-shaped surface region.
It is parameterized by a vector $\overrightarrow{\bm{o_{\mathbb{H}}}\bm{p_{i}}}\coloneqq(\upsilon_{x},\upsilon_{y})$ that is located at the palm center $o_{\mathbb{H}}$, i.e., the origin of $\{\mathbb{H}\}$, and pointing to the contact point $\bm{p}_{i}$.
The value ranges of $\upsilon_{x}$ and $\upsilon_{y}$ restrict $\bm{p}_{i}$ such that it can only move within the palm's inner surface.
\section{Calculation of approximated friction cone}
\label{app:friction_cone}
The friction cone is approximated by an ${N}_{f}$-sided pyramid:
\begin{equation}
    \mathcal{F} = \begin{bmatrix}
           & \mu \cos(\frac{2\pi i}{n}) &       \\
    \cdots & 1                          & \cdots\\
           & \mu \sin(\frac{2\pi i}{n}) &
    \end{bmatrix}_{3\times {N}_{f}},~n=1,2,\dots,{N}_{f}.
\end{equation}
Each column represents one edge of the approximated friction cone, denoted as $\mathbf{f}_{n}$, and is normalized such that $\|\mathbf{f}_{n}\|=1$.
This assumes that all contact forces have the same limit.
The position of entry $1$ in $\mathbf{f}_{n}$ indicates that the central axis of the pyramid coincides with the corresponding axis of $\{\mathbb{C}\}$.
\end{appendices}
%

%
\vspace{-33pt}
\begin{IEEEbiography}[{\includegraphics[width=1in,height=1.25in,clip,keepaspectratio]{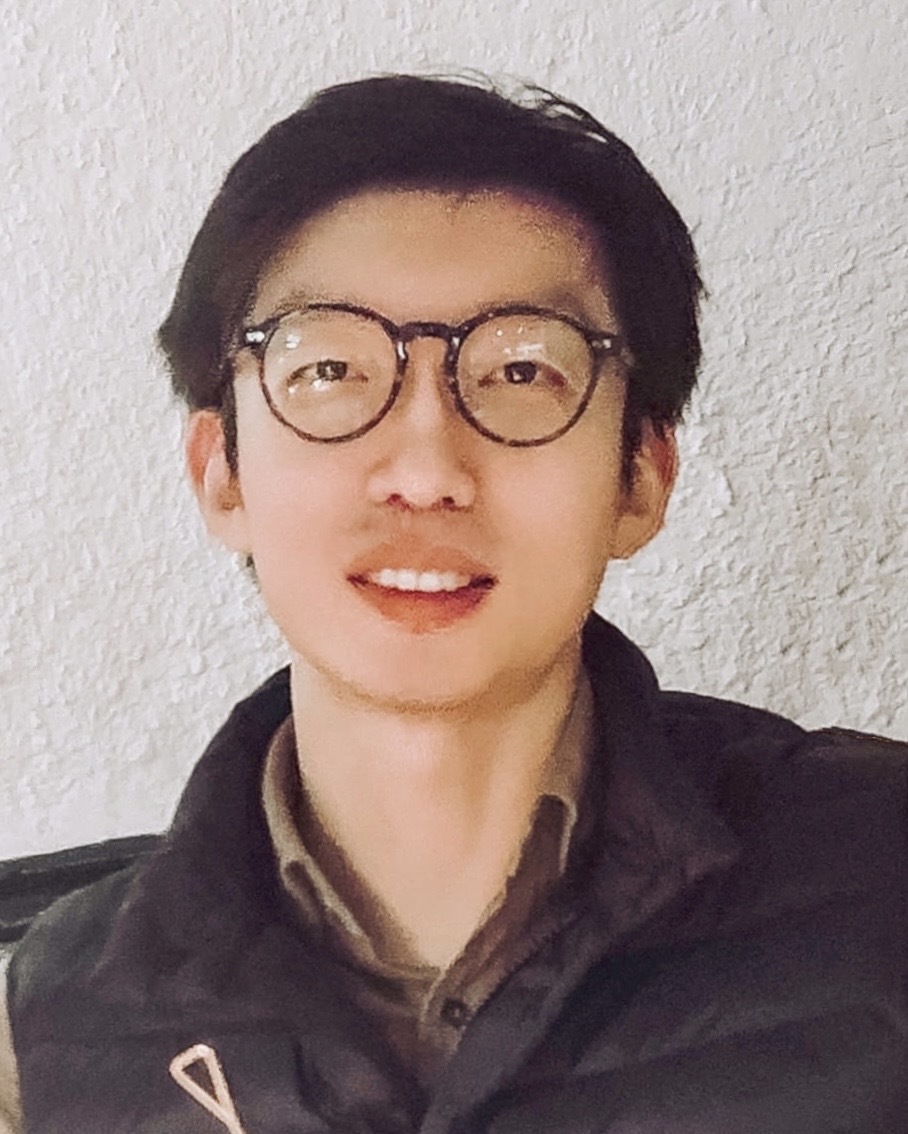}}]{Kunpeng Yao} (Member, IEEE) received a B.Sc. degree from Shanghai Jiao Tong University (SJTU), China, in 2013; an M.Sc. degree from the Technical University of Munich (TUM), Munich, Germany, in 2017; and his Ph.D. degree from the Swiss Federal Institute of Technology Lausanne (EPFL), Lausanne, Switzerland, in 2022.
He is currently a postdoctoral researcher at LASA, EPFL.
His main research interests include human motor control, robotic grasping, and dexterous manipulation.
\end{IEEEbiography}
\vspace{-33pt}
\begin{IEEEbiography}
[{\includegraphics[width=1in,height=1.25in,clip,keepaspectratio]{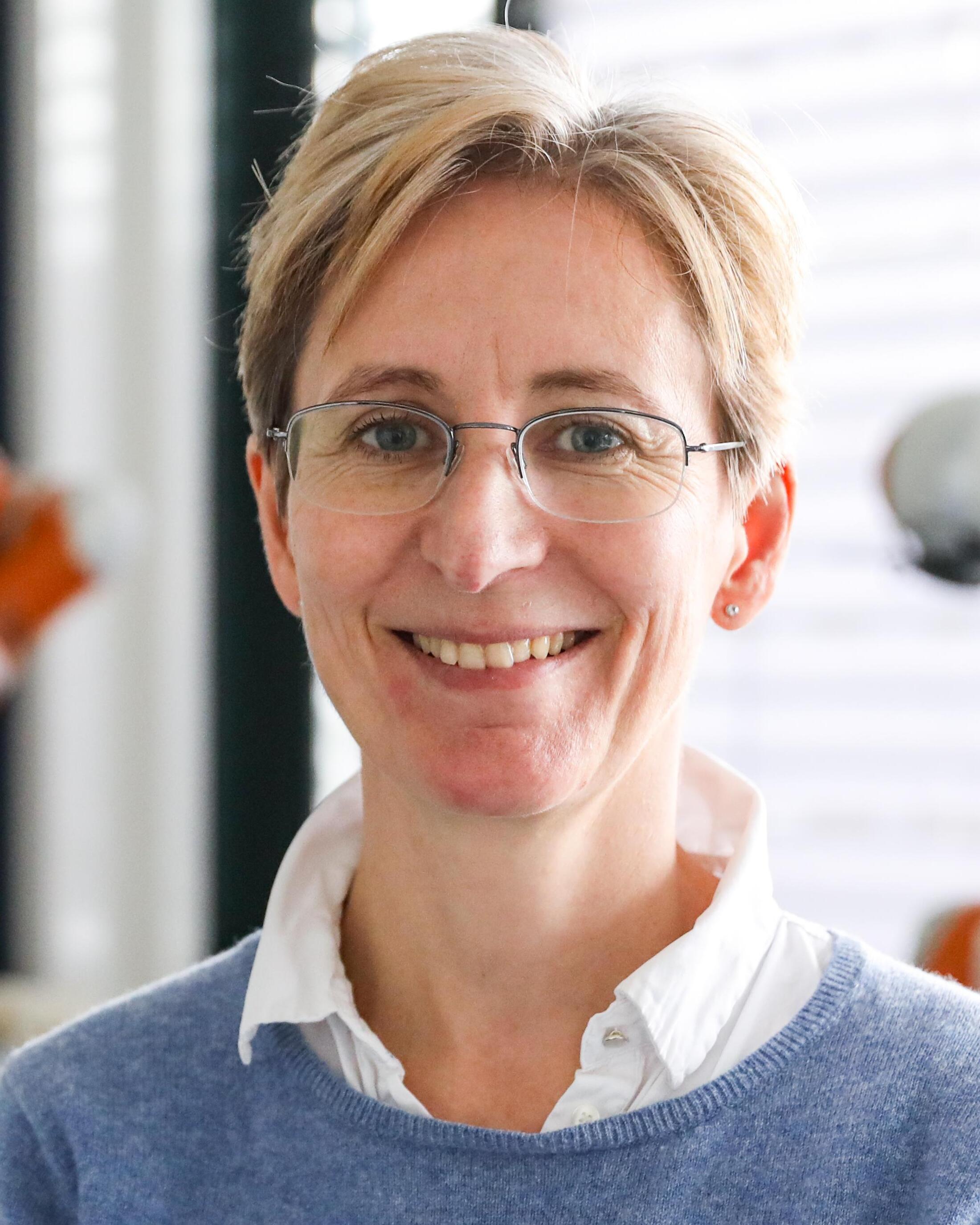}}]{Aude Billard} (Fellow, IEEE) received an M.Sc. degree in physics from the Swiss Federal Institute of Technology Lausanne (EPFL), Lausanne, Switzerland, in 1995, and an M.Sc. degree in knowledge-based systems, and her Ph.D. degree in artificial intelligence from the University of Edinburgh, Edinburgh, U.K., in 1996 and 1998, respectively.
She is currently a Full Professor of Micro and Mechanical Engineering and the head of the Learning Algorithms and Systems Laboratory, School of Engineering, EPFL.
Her research spans the fields of machine learning and robotics, including modelling dexterity and fast adaptive control to enable skill acquisition in robots and applying dynamical systems approach to physical human-robot interactions.
\end{IEEEbiography}
\end{document}